\documentclass[journal=jacsat,manuscript=article]{achemso}

\usepackage[version=3]{mhchem} 
\usepackage{microtype}
\usepackage{hyperref}
\usepackage{url}
\usepackage{booktabs}

\usepackage{amsmath}
\usepackage{amssymb}
\usepackage{mathtools}
\usepackage{amsthm}
\usepackage{threeparttable}
\usepackage{pifont}
\usepackage{xcolor}
\usepackage{makecell}
\usepackage{listings}
\usepackage{graphicx}
\usepackage{caption}
\usepackage{lipsum}
\usepackage{multirow}

\newcommand{\greencheck}{\textcolor{green}{\ding{52}}}
\newcommand{\redcross}{\textcolor{red}{\ding{56}}}

\usepackage{lineno}

\author{Xingyu Dang}
\altaffiliation{These authors contributed equally to this work.}
\affiliation[Princeton]
{Department of Computer Science, Princeton University, Princeton, NJ, 08540, United States}
\author{Haocheng Tang}
\altaffiliation{These authors contributed equally to this work.}
\affiliation[University of Pittsburgh]
{School of Pharmacy, University of Pittsburgh, Pittsburgh, PA, 15261, United States}
\alsoaffiliation[Northeastern University]
{Khoury College of Computer Science, Northeastern University, Boston, MA, 02115, United States}
\author{Junmei, Wang}
\affiliation[University of Pittsburgh]
{School of Pharmacy, University of Pittsburgh, Pittsburgh, PA, 15261, United States}
\email{juw79@pitt.edu}
\author{Yanjun, Li}
\affiliation[UF-MedChem]
{Department of Medicinal Chemistry, Center for Natural Products, Drug Discovery and Development, University of Florida, Gainesville, FL, 32610, United States}
\alsoaffiliation[UF-CISE]
{Department of Computer and Information Science and Engineering, University of Florida, Gainesville, FL, 32611, United States}
\email{yanjun.li@ufl.edu}

\title[An \textsf{achemso} demo]
  {Learning Mechanistic Reasoning for Chemical Reactions with Large Language Models}

\abbreviations{ML, LLM, Mechanism, Organic chemistry}

\begin{document}
\begin{tocentry}

\centering
\includegraphics[width=9cm,height=3.5cm,keepaspectratio]{figures/Figure1.pdf}

\end{tocentry}

\begin{abstract}
  Reaction mechanisms consist of the step-by-step sequences of elementary reactions that explain chemical transformations. Learning the mechanism logic is therefore essential for enhancing the fundamental chemical intelligence of large language models (LLMs). The stepwise deduction of reaction mechanism aligns naturally with the reasoning paradigms of reasoning LLMs. However, current chemical LLMs primarily emphasize coarse-grained name reactions for product prediction and retrosynthesis, often leading to physical inconsistencies and hallucinations. In contrast, specialized small-scale generative models for mechanism inference typically suffer from restricted generalization capacity across diverse chemical spaces. To overcome these limitations, we built a novel, large-scale reasoning dataset of reaction mechanisms. Furthermore, we established the FukuyamaBench, a difficult benchmark derived from Fukuyama’s Advanced Organic Reaction Mechanism book, to rigorously evaluate model performance on hierarchical mechanism reasoning. Our fine-tuned Qwen3-30B-A3B achieves 8.3\% exact pathway match on FukuyamaBench Set~A, surpassing the specialized FlowER model (5.1\%), demonstrating that mechanism-aware training substantially enhances chemical reasoning in language models.
\end{abstract}

\section{Introduction}

The fundamental role of reaction mechanisms in organic chemistry cannot be overstated; they are the step-by-step sequences of elementary reactions that describe precisely how chemical changes occur at the molecular level \citep{smith2006march, flynn2015mechanisms, grossman2019art}. Traditionally, predicting these transformations has relied heavily on recognizing coarse-grained reaction types, also known as ``name reactions'', which act as convenient but superficial labels \citep{kurti2005strategic}. While this label-dependent approach is common in current chemical Large Language Models (LLMs) for product prediction and retrosynthesis, it inherently lacks true chemical insight. By prioritizing overall transformations over fundamental reactivity, these models frequently struggle with out-of-domain cases, leading to physical inconsistencies and chemical hallucinations \citep{bradshaw2025review}. Conversely, specialized small-scale generative models designed for mechanism inference typically suffer from restricted generalizability across diverse chemical spaces. We argue that achieving authentic chemical intelligence requires AI systems to ground their reasoning in intrinsic chemical reactivity rather than relying on memorized, label-like name reactions. This mechanistic perspective is particularly important for laboratory synthesis, where chemically plausible reasoning is necessary to handle nuanced and context-sensitive reaction scenarios.

The step-by-step deduction of a reaction mechanism is intrinsically a Markov-like process, where each intermediate state dictates the next logical elementary step based on localized electron flow and steric environments. This sequential logic aligns naturally with the Chain-of-Thought (CoT) reasoning paradigms of modern reasoning LLMs. Compared to small generation models, LLMs offer distinct advantages for this task. First, they allow for the natural injection of complex reaction conditions—such as temperature, light condition, equivalent weight and catalysts—directly into the reasoning context, enabling the prediction of condition-responsive reactions (fig.~\ref{fig:1}a). Second, reasoning LLMs possess the contextual capacity to map out divergent pathways, predicting major products while simultaneously rationalizing the formation of minor byproducts in reactions with multi-products. Most importantly, an LLM equipped with mechanism-aware reasoning can successfully navigate complex reactions that extend far beyond simple name reactions, significantly enhancing its generalizability to out-of-domain chemical spaces (fig.~\ref{fig:1}b). However, constructing high-quality mechanistic training data poses a unique challenge: the annotator must have access to the ground-truth reaction pathway to ensure chemical correctness, yet the resulting reasoning must read as forward prediction so the student model learns to deduce---not merely recall---each step. Based on these insights, the main contributions of this article lie in the following three aspects:

\begin{figure}
    \centering
    \includegraphics[width=1\linewidth]{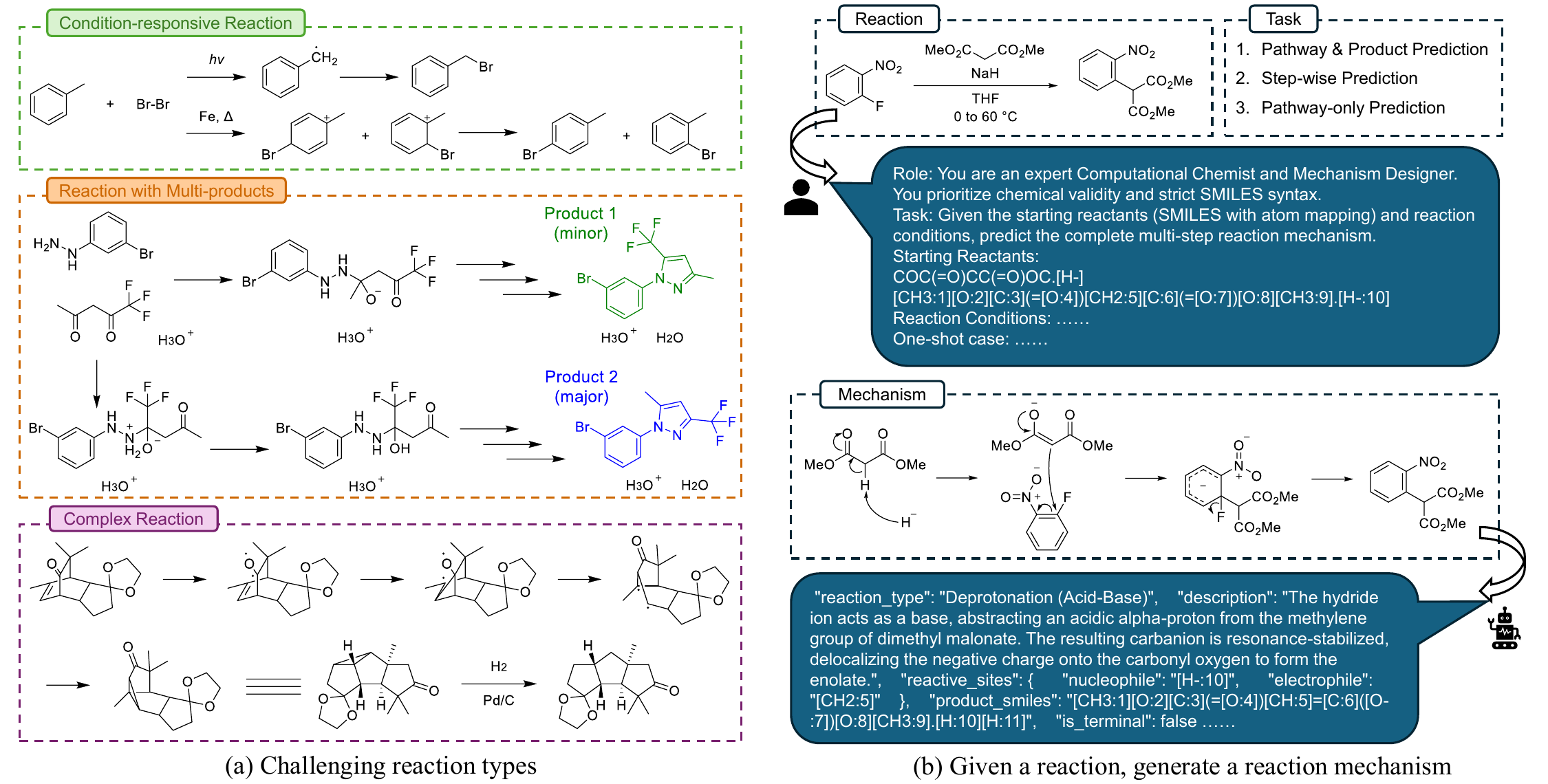}
    \caption{From Complex Reaction Challenges to Mechanistic Reasoning with LLMs.}
    \label{fig:1}
\end{figure}

\begin{itemize}
    \item \textbf{Establishment of FukuyamaBench:} To rigorously evaluate hierarchical mechanism reasoning, we established FukuyamaBench, derived from Fukuyama’s renowned workbook on organic reaction mechanisms. Spanning real-world problems from graduate exams to cutting-edge research, this multi-tiered benchmark comprises 1,997 intermediate steps across diverse fundamental reaction types. Evaluations on FukuyamaBench demonstrates that mechanism-aware post-training effectively transforms LLMs from simple pattern-matchers towards potential chemical reasoners.

    \item \textbf{Large-Scale Reasoning Dataset for Mechanisms:} We built a novel, large-scale reasoning dataset of reaction mechanisms for LLM post-training. Each training example contains structured, forward-predictive reasoning followed by atom-mapped intermediate SMILES at every elementary step. By deducing mechanisms step-by-step, the model internalizes fundamental rules like electron flow and intermediate stability, ensuring predictions are grounded in physical reality rather than coarse-grained name-reaction labels.
    
    \item \textbf{Empirical Validation via Training \texttt{Qwen3-30B-A3B}:} We validated our approach using Supervised Fine-Tuning (SFT) and customized Reinforcement Learning (RL) on the \texttt{Qwen3-30B-A3B-Instruct} model \citep{yang2025qwen3technicalreport}. Evaluations against the Flower validation set and our benchmark demonstrate that mechanism-aware training significantly enhances the model’s comprehension, transforming it from a superficial pattern-matcher into a reasoner capable of tackling difficult real-world problems.
\end{itemize}

In summary, this work bridges the gap between superficial pattern matching and genuine mechanistic reasoning. By introducing FukuyamaBench and a chemically grounded training pipeline, we provide both rigorous evaluation criteria and an effective post-training recipe for chemical LLMs. Our results show that mechanism-aware post-training substantially improves chemical reasoning in a 30B MoE model, though a significant gap to frontier models remains, suggesting that scaling both data coverage and model capacity is a promising direction.

\section{Related Work}

\subsection{LLMs for Chemistry}

LLMs have been extensively explored in chemistry, covering molecular understanding, reaction reasoning, and scientific knowledge grounding. Early efforts primarily focused on prompt engineering, demonstrating that general-purpose LLMs can solve chemistry tasks through carefully designed prompts without parameter updates \citep{Hatakeyama-Sato2023pegpt4, Liu2024pe}. Building upon these observations, a series of chemistry-specific fine-tuned models have been proposed, including ChemLLM \citep{zhang2024chemllm}, ChemDFM \citep{zhao2025developing}, and BatGPT-Chem \citep{Yang2025batgptchem}, which adapt large pretrained backbones to domain-specific corpora and tasks.

To overcome the limitations of text-only representations, multimodal LLMs for chemistry have incorporated structured molecular modalities, such as SMILES, graphs, and 3D geometries. Representative models include nach0 \citep{Livne2024nach0}, InstructMol \citep{cao2023instructmol}, Chem3DLLM \citep{jiang2025chem3dllm}, and ChemVLM \citep{li2025chemvlm}, enabling joint reasoning over language and molecular structures. More recently, reasoning-oriented chemical LLMs have emerged, leveraging reinforcement learning or structured supervision to enhance multi-step chemical reasoning and interpretability, exemplified by ether0 \citep{narayanan2025ether0}, ChemDFM-R \citep{zhao2025chemdfmr}, and Chem-R \citep{wang2025chemr}.

Despite these advances, fine-grained reasoning over reaction pathways remains largely unexplored from an organic chemistry perspective. Most existing approaches focus on the reaction-level predictions, whereas the step-wise structural transformations and mechanistic pathways that are central to organic chemistry remain insufficiently modeled.

\subsection{Generative Models in Organic Reaction Mechanisms}

Beyond reaction-level prediction, several works have explored generative modeling of organic reaction mechanisms and step-wise molecular transformations. Sequence-based approaches such as T5Chem \citep{lu2022t5chem} and PMechRP \citep{miller2025pmechrp} model reaction pathways using text or symbolic representations, demonstrating the feasibility of learning mechanistic knowledge from annotated data. In parallel, structure-aware models explicitly operate on molecular graphs or electron flows. Graph2SMILES (G2S) \citep{joung2024graph2smiles} formulates mechanisms as sequences of graph edits, while FlowER \citep{joung2025flower} introduces electron flow matching to generate elementary reaction steps in a physically motivated manner.

More recently, benchmarks such as ChemCoTBench \citep{li2025chemcotbench} and oMeBench \citep{xu2025omebench} have been proposed to evaluate step-wise chemical reasoning, with an emphasis on intermediates and pathway consistency. Despite these advances, current datasets and models often underemphasize the fundamental role of reaction mechanisms in organic chemistry, including understanding how reactions proceed, predicting major products, and rationalizing byproducts and selectivity observed in experiments.

\begin{figure}[t]
    \centering

    \begin{minipage}{\textwidth}
        \centering
        \captionof{table}{Comparison of representative organic reaction mechanism inference benchmarks}
        \label{tab:1}
        \resizebox{\textwidth}{!}{
        \begin{tabular}{lcccccccc}
            \toprule
            Benchmark & Reactions & Steps & Level & Task & Int.? & Cons.? & Map.? & Beyond Name?\\
            \midrule
            ChemCoTBench & 85 & 85 & Easy & \makecell{Step-wise \\ Selection} & \redcross & \redcross & \redcross & \redcross\\
            oMe-Gold & 196 & 858 & Mixed & Pathway & \greencheck & \redcross & \redcross & \redcross \\
            FukuyamaBench & 319 & 1997 & Mixed & \makecell{Pathway \\ Step-wise \\ Product} & \greencheck & \greencheck & \greencheck & \greencheck\\
            \bottomrule
        \end{tabular}
        }
    \raggedright
    \scriptsize
    ``Int.?'', ``Cons.?'', and ``Map.?'' indicate whether the benchmark considers the mechanistic intermediates, conservation of atomic and electronic at each elementary step, and skeleton atom mapping, respectively. ``Beyond Name?'' indicates whether the benchmark includes complex reactions beyond simple, predefined name reactions. 
    \end{minipage}

    \vspace{1em}

    \begin{minipage}{\textwidth}
        \centering
        \includegraphics[width=\linewidth]{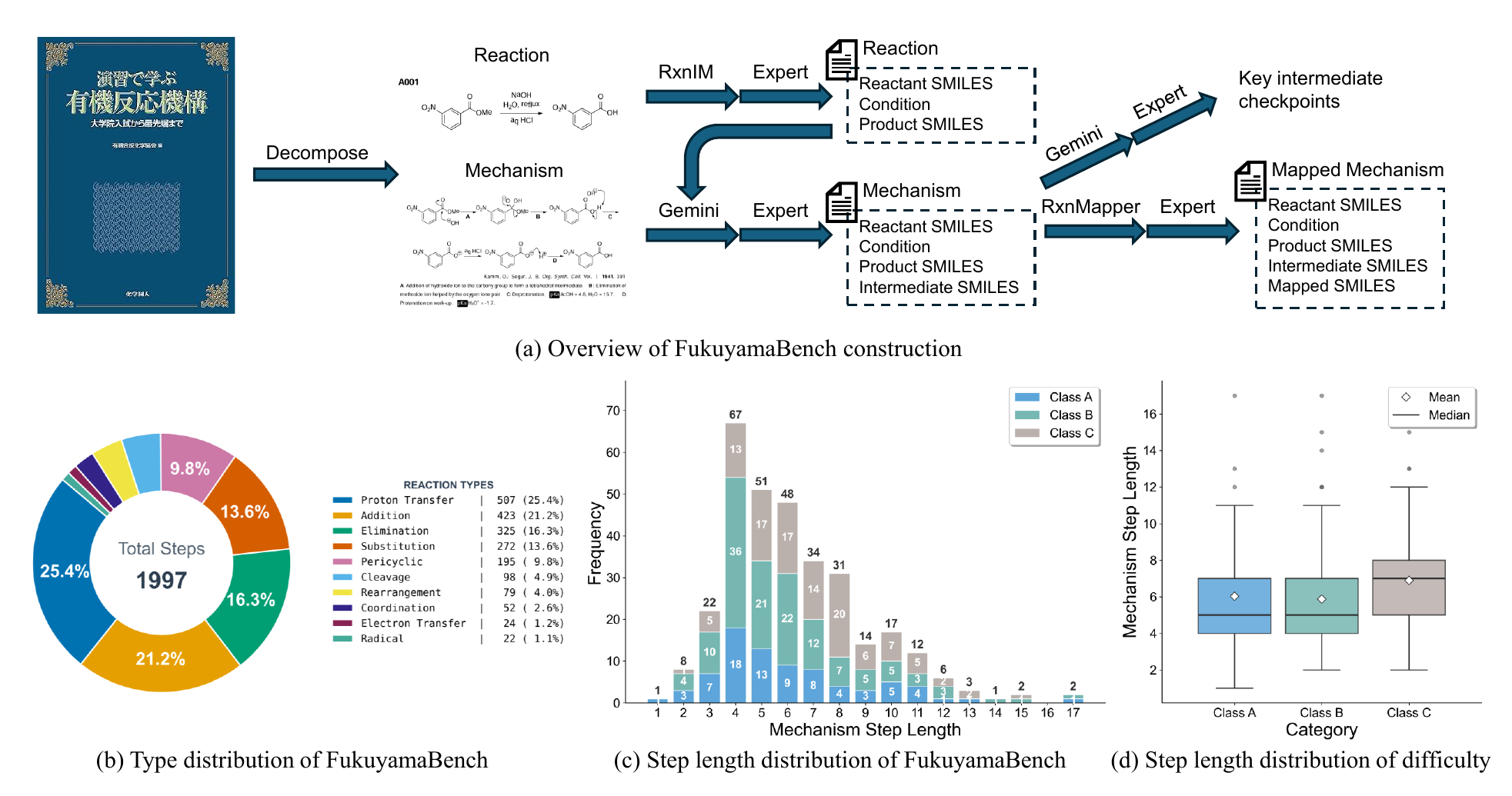}
        \caption{Overview of FukuyamaBench construction and metadata}
        \label{fig:2}
    \end{minipage}

\end{figure}

\section{Methodology}

\subsection{FukuyamaBench}

Tab.~\ref{tab:1} highlights four key limitations of previous benchmark: (i) \textbf{single-task}. Early work focused only on one of the three aspects of reaction mechanisms: predicting products, intermediates, or explaining the reasons for the occurrence of reactions. ; (ii) \textbf{non-conservation}. The lack of step-wise atomic conservation and electronic conservation does not conform to the most fundamental principles of chemical reactions.; (iii) \textbf{lack of atom mapping}. The simple SMILES representation lacks the ability to perceive the relative positions of the skeleton and the reaction site.; (iv) \textbf{reliance on name reactions}. Existing datasets are largely confined to simple, predefined name reactions, failing to evaluate a model's capacity to navigate complex, condition-responsive transformations based on intrinsic chemical reactivity. Furthermore, when we examined previous benchmarks such as ChemCoTBench, we found that many of its SMILES representations were incorrect. For instance, in the structure \texttt{c1ccc(\allowbreak[PH](c2ccccc2)(c2ccccc2)\allowbreak[Pd](\allowbreak[PH](c2ccccc2)(c2ccccc2)c2ccccc2)\allowbreak[PH](c2ccccc2)(c2ccccc2)c2ccccc2)cc1}, there was an excessive \texttt{H} atom present on the \texttt{P} atom, which further reduced the credibility of these benchmarks.

To address these limitations, we introduced FukuyamaBench, a comprehensive organic reaction mechanism benchmark based on the \textit{Advanced Organic Reaction Mechanism} workbook \citep{Fukuyama2005Mechanism}, a classic collection of organic chemistry problems curated by the Society of Synthetic Organic Chemistry, Japan. This book consists of 3 question sets: A, B, C and an answer section. Set A contains fundamental and crucial reaction questions, while set B includes questions at the graduate level or of slightly higher difficulty, and set C comprises questions collected from researchers in real labs. Set A and Set B are also regarded as classic textbooks for the Chemistry Olympiad.

\textbf{Construction}. 
First, the problem sets were decomposed into two image categories: question-reactions and answer-mechanisms. For the reaction images, RxnIM was employed to extract reactants, conditions and products. The extracted outputs were subsequently reviewed and manually corrected by domain experts to generate clean reaction texts. These reaction texts along with the mechanism image set, were then input into \texttt{Gemini-3-Pro}, and the obtained mechanism texts were validated by domain experts. Subsequently, the verified mechanism texts were combined with predefined checkpoint rules and processed through \texttt{Gemini-3-Pro} once more. Experts corrected and confirmed the results for reaction intermediates, ensuring the step-wise atomic and electronic conservation by RDKit. Finally, the reaction equations were submitted to RxnMapper \citep{schwaller2021extraction}, and the corresponding mapped RXN SMILES were verified by experts through generated pictures. The overview workflow is shown in fig.~\ref{fig:2}a.

\textbf{Multi-tasks}.
Given a reaction mechanism trajectory \( T \), it consists of a sequence of intermediate--condition pairs $\{ (x_0 \mid c_0), (x_1 \mid c_0), \ldots, (x_i \mid c_0), (x_i \mid c_1), \ldots, (x_n \mid c_m) \} \subseteq T.$, where $x_p$ stands for intermediate and $c_q$ represents the corresponding reaction condition. We then defined three tasks: \emph{Step-wise Only Prediction}, \emph{Pathway-only Prediction}, and \emph{Pathway \& Product Prediction}: 

\begin{itemize}
    \item \textbf{Step-wise Only Prediction}: given an intermediate--condition pair \( (x_j \mid c_k) \), the objective is to infer the subsequent intermediate \( (x_{j+1} \mid c_k) \).
    \item \textbf{Pathway-only Prediction}: given the initial intermediate--condition pair \( (x_0 \mid c_0) \), the sequence of conditions \( c_0, \ldots, c_m \), and the final intermediate--condition pair \( (x_n \mid c_m) \), the task is to infer the intermediates \( x_1, \ldots, x_{n-1} \).
    \item \textbf{Pathway \& Product Prediction}: given the initial intermediate--condition pair \( (x_0 \mid c_0) \) and the sequence of conditions \( c_0, \ldots, c_m \), the task is to jointly infer the remaining intermediates and products \( x_1, \ldots, x_n \).
\end{itemize}

\textbf{Coarse-grained Checkpoint}.
Two main challenges arise in evaluating on FukuyamaBench: (i) Some problems in FukuyamaBench are coarse-grained. Scientists often overlook obvious or repetitive steps when writing the mechanism. (ii) Unlike the step-wise generative model FlowER, LLMs tend to generate coarse-grained steps. Thus, the alignment method proposed by oMeS is not suitable for FukuyamaBench, since oMe-Gold is a fine-grained dataset. Here, we established a rule-based method for generating intermediate checkpoints: (i) change in heavy atom connectivity or bond topology. Any product resulting from steps involving the formation or breakage of C-C, C-N, C-O or C-X bonds must be regarded as a checkpoint; (ii) transformation of functional groups. When a molecule changes from one type of compound to another, its representative structure should be recorded; (iii) Exclusion of trivial proton transfers. In the absence of changes in the acid-base conditions, consecutive proton gain and loss steps should be combined into the same checkpoint.

\textbf{Conservation}.
In each step of the mechanism, the reactants and products satisfy the laws of atomic conservation and electron conservation.

\textbf{Mapping}.
The reactants and products within each step of the mechanism ensure that the backbone atoms correspond accordingly. The numbers of the atoms at which the reaction occurs can be clearly seen.

\subsection{Training Dataset Construction}

 \begin{figure}
    \centering
    \includegraphics[width=0.95\linewidth]{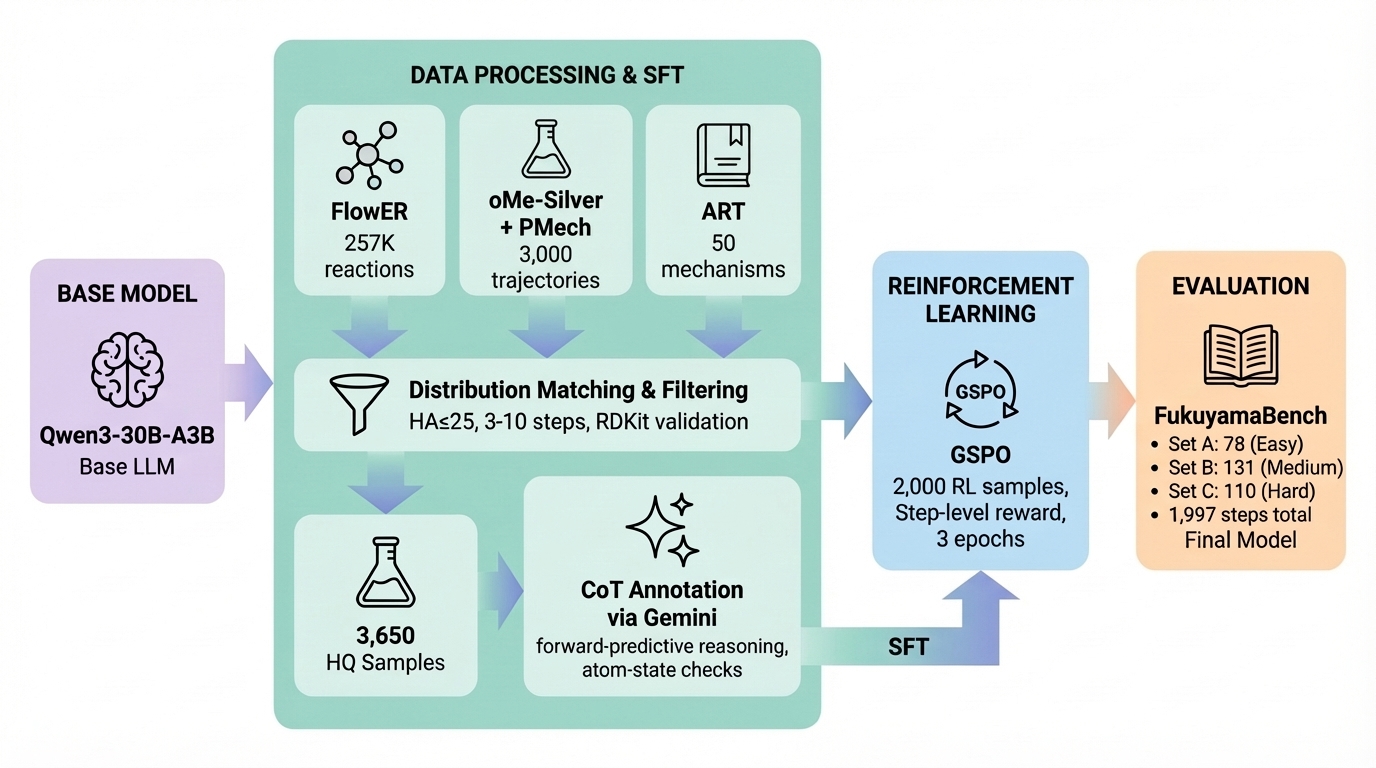}
    \caption{The overall training pipeline}
    \label{fig:3}
\end{figure}

The training data is constructed from four primary sources (fig.~\ref{fig:3}): (1)~FlowER \citep{joung2025flower}, a large-scale database of 257K elementary reaction steps; (2)~oMe-Silver \citep{xu2025omebench} and PMech \citep{miller2025pmechrp}, which together provide 3,039 reaction trajectories with explicit conditions (reagents, solvents, temperature); (3)~\textit{The Art of Writing Reasonable Organic Reaction Mechanisms} \citep{grossman2019art}, from which 50 complex mechanisms are manually extracted; and (4)~Gemini-synthesized reactions for targeted coverage of underrepresented reaction families.

\textbf{Quality filtering.} FlowER's raw 257K reactions are predominantly large pharmaceutical molecules (median 42 heavy atoms, 4.6 rings), reflecting its origins in industrial synthesis databases. For such large substrates, the reaction mechanisms tend to be excessively long and repetitive, making it difficult for the model to learn the underlying mechanistic principles. We therefore apply quality-driven filtering to focus on cleaner, more interpretable reactions: we retain reactions with moderate molecular size, 3--10 elementary steps, valid SMILES, and no duplicates. This yields 2,353 high-quality FlowER samples that emphasize core organic reaction types (nucleophilic substitution, elimination, addition, rearrangement) where electron flow and competing pathways can be clearly analyzed. Combined with 1,297 condition-annotated external samples, this produces 3,650 SFT training examples. For RL, we use 2,000 reactions (1,098 external + 902 FlowER) filtered by similar criteria as the on-policy training set.

\textbf{Reaction type augmentation.} FlowER, being derived from patent and pharmaceutical literature, has limited coverage of several reaction families that are central to organic chemistry textbooks, notably pericyclic reactions (Diels-Alder, electrocyclization, sigmatropic rearrangements), which constitute $<$0.1\% of FlowER but are fundamental to any comprehensive organic chemistry curriculum. To broaden the training distribution, we prompt \texttt{Gemini-3.0-Pro} to generate atom-mapped mechanisms for 18 pericyclic/radical families (882 reactions) and 16 named reaction families (124 reactions, including Wittig, Beckmann, Suzuki, Swern, Fischer indole, etc.). Each generated mechanism follows an explicit textbook template to ensure chemical accuracy, and all SMILES are validated with RDKit. To avoid data contamination, we never use any benchmark problem as a seed---instead, Gemini creates novel substrates from prompted reaction types. All training data is fully disjoint from the evaluation benchmark.

\textbf{CoT annotation.} A key design challenge is that the annotator (expert model) has access to the full ground-truth trajectory---starting reactants, all intermediates, and final products---yet must produce reasoning that reads as \emph{forward prediction}, as if the answer were unknown. This is essential because the student model at inference time sees only the starting reactants.

To achieve this, we adopt a \textbf{teacher--student decoupling} strategy. The \emph{annotator prompt} (teacher) receives the complete reaction trajectory as context and is instructed to write in predictive language (``the nucleophile \emph{will} attack'', ``the intermediate \emph{is expected to} rearrange'') while explicitly forbidding hindsight phrases (``as shown in the product'', ``the result confirms''). The annotator copies product SMILES exactly from the ground truth, ensuring chemical correctness, but frames the reasoning as a forward deduction. During \emph{conversion to training data}, the ground-truth trajectory is stripped from the user prompt, so the student model only sees starting reactants and conditions---never the answer it must predict.

Several annotation design choices further improve data quality: (i)~\textbf{Atom ID audit}: the annotator must verify that all atom map IDs referenced in the reasoning actually exist in the current step's reactant SMILES, preventing hallucinated atom references; (ii)~\textbf{Adaptive granularity}: obvious steps (e.g., proton transfer to a strong base) receive concise 1--2 sentence explanations, while decision points involving selectivity, competing pathways, or rearrangements receive detailed multi-sentence analysis; (iii)~\textbf{Structured output}: each step produces a JSON object with reaction type, reactive sites (nucleophile, electrophile, leaving group), and product SMILES, enabling automated validation. All annotations are validated with RDKit for SMILES correctness, and malformed entries are filtered before training.

\subsection{Training Pipeline}

\textbf{Supervised Fine-tuning.}
We perform fine-tuning of \texttt{Qwen3-30B-A3B-Instruct} on 3,650 scope-matched samples for 3 epochs. Each training example follows a structured four-phase reasoning template: (A)~\emph{Mechanism Analysis}---analyze electron flow and identify bond changes; (B)~\emph{Atom State Definition}---explicitly define charge and valence of reaction centers before generating SMILES; (C)~\emph{SMILES Syntax Translation}---apply rule checks (e.g., no pentavalent carbons); (D)~\emph{Predicted Intermediate}---output the final SMILES. This structured reasoning enforces chemical validity at each step. Training details are in Appendix~\ref{app:training}.

We note that SFT alone underperforms the base model on exact pathway match. We found that the SFT model learns correct chemistry---its checkpoint accuracy is high (59\%)---but tends to over-generate intermediate steps, making exact pathway match harder. This is consistent with prior findings that distilling reasoning traces from large teacher models into smaller ones produces excessively verbose outputs\citep{zhang2025making}. In our setting, the reasoning annotations generated by Gemini contain detailed multi-phase analysis that the student model internalizes as lengthy outputs, inflating predicted step counts beyond what exact match requires.

\textbf{Reinforcement Learning.}
During RL, we optimize the policy with Group Sequence Policy Optimization (GSPO) \citep{zheng2025groupsequencepolicyoptimization}, which we prefer over GRPO \citep{shao2024deepseekmathpushinglimitsmathematical} for two reasons: (i)~sequence-level grouping and clipping reduces reward variance in long-horizon mechanism prediction, and (ii)~it avoids routing replay overhead, making it practical for MoE models. Hyperparameters are detailed in Appendix~\ref{app:training}.

We use a \textbf{step-level reward design}. The reward is computed at reaction-step level and averaged over the full mechanism trajectory. Let the ground-truth product for step $t$ be $g_t$ and the predicted product be $\hat{g}_t$. After RDKit canonicalization (including atom-map removal and component sorting), each step receives
\[
s_t=
\begin{cases}
1.0, & \text{if } \hat{g}_t \equiv g_t,\\
0.1, & \text{if } \hat{g}_t \not\equiv g_t \text{ but is a valid SMILES},\\
0.0, & \text{otherwise}.
\end{cases}
\]
The final trajectory reward is $r=\frac{1}{T}\sum_{t=1}^{T} s_t$, where missing predicted steps are scored as $0.0$. To enforce output format reliability, we apply structured penalties before step scoring: if the response does not contain the required ``\#\# Result'' section, the reward is set to $-0.3$; if the section exists but JSON parsing fails, the reward is set to $-0.2$. These penalties reduce malformed generations and make downstream scoring deterministic.

\section{Experiment}

\subsection{Performance on FlowER Test Dataset}

To verify that mechanism-aware training generalizes beyond the benchmark, we evaluate on the FlowER validation set (500 held-out reactions). The base \texttt{Qwen3-30B-A3B} achieves 0.0\% step-wise accuracy with only 28.7\% SMILES validity, indicating that the pretrained model cannot produce chemically valid mechanism intermediates without domain-specific training. After the training pipeline, our model reaches 21.5\% step-wise accuracy and 88.6\% SMILES validity which confirms the model has internalized generalizable mechanistic knowledge rather than memorizing specific patterns from training set.

\subsection{Performance on FukuyamaBench}

\begin{table}[htbp]
    \centering
    \caption{The result of pathway \& product prediction task}
    \label{tab:2}
    \resizebox{\linewidth}{!}{
    \begin{tabular}{lcccccccccccc}
        \toprule
        & \multicolumn{4}{c}{\textbf{A}} & \multicolumn{4}{c}{\textbf{B}} & \multicolumn{4}{c}{\textbf{C}} \\
        \cmidrule(lr){2-5} \cmidrule(lr){6-9} \cmidrule(lr){10-13}
        & Top-1 & Top-3 & Top-5 & Top-8 & Top-1 & Top-3 & Top-5 & Top-8 & Top-1 & Top-3 & Top-5 & Top-8 \\
        \midrule
        FlowER \citep{joung2025flower} & 5.1\% & 6.4\% & 6.4\% & 6.4\% & 0.8\% & 1.5\% & 2.3\% & 2.3\% & 0.0\% & 0.0\% & 0.0\% & 0.0\% \\
        GPT-5 & 10.0\% & 12.9\% & 15.5\% & 19.2\% & 4.3\% & 5.2\% & 5.9\% & 6.9\% & 0.9\% & 0.9\% & 0.9\% & 0.9\% \\
        Gemini-3.0-Pro & 22.3\% & 31.7\% & 36.0\% & 39.7\% & 7.8\% & 10.7\% & 12.3\% & 13.7\% & 1.0\% & 1.2\% & 1.5\% & 1.8\% \\
        Qwen3-235B-A22B & 9.8\% & 15.4\% & 18.2\% & 20.5\% & 4.0\% & 4.7\% & 5.1\% & 5.3\% & 0.2\% & 0.6\% & 0.8\% & 0.9\% \\
        Qwen3-30B-A3B & 6.9\% & 8.6\% & 9.0\% & 9.0\% & 3.8\% & 4.5\% & 4.6\% & 4.6\% & 0.9\% & 0.9\% & 0.9\% & 0.9\% \\
        \midrule
        ChemDFM-R-14B \citep{zhao2025chemdfmr} & 5.8\% & 6.4\% & 6.4\% & 6.4\% & 3.4\% & 4.1\% & 4.4\% & 4.6\% & 0.6\% & 0.9\% & 0.9\% & 0.9\% \\
        \midrule
        Ours & \textbf{8.3\%} & \textbf{12.1\%} & \textbf{13.9\%} & \textbf{15.4\%} & 2.8\% & 3.3\% & 3.5\% & 3.8\% & 0.0\% & 0.0\% & 0.0\% & 0.0\% \\
        \bottomrule
    \end{tabular}
    }
\end{table}


To evaluate the chemical reasoning and multi-step reasoning capabilities of our proposed framework, we conducted extensive experiments on the FukuyamaBench. As described in the previous section, this benchmark requires models to perform joint prediction of reaction pathways and final products, which represents a significant challenge even for the most advanced LLMs. We evaluated the performance of models on all three tasks: \emph{Step-wise Only Prediction}, \emph{Pathway-only Prediction}, and \emph{Pathway \& Product Prediction}.

\textbf{Evaluation metric.}
For each problem, we draw $n$ independent mechanism trajectories from the model and report \emph{pass@k} ($k\!\in\!\{1,3,5,8\}$), the probability that at least one of $k$ randomly chosen samples is an exact pathway match, computed with the unbiased estimator $\text{pass@}k = 1 - \binom{n-c}{k}/\binom{n}{k}$ \citep{brown2024large, dang2025weight}.
In mechanism prediction, pass@$k$ captures \emph{predictive diversity}: a single correct trajectory among $k$ attempts suffices for a chemist to identify the plausible pathway. Note that unlike FlowER's conditioned beam search, LLM samples are drawn independently, making Top-$k$ a strictly harder metric---yet LLMs consistently outperform FlowER on FukuyamaBench.

The results for the pathway and product prediction task are summarized in Tab.~\ref{tab:2}. We observed that general-purpose models, while possessing strong linguistic reasoning, often struggle with the precise structural requirements of complex chemical syntheses. For instance, in Category A, \texttt{GPT-5} and \texttt{Gemini-3.0-Pro} achieved Top-8 accuracies of 19.2\% and 39.7\%, respectively. The expert model in the field of chemistry, such as \texttt{ChemDFM-R-14B}, which has undergone domain-specific fine-tuning, also exhibited poor performance due to insufficient training on reaction mechanisms.

\begin{table}[htbp]
\centering
\caption{Pathway-only prediction (Top-1 accuracy). The model is given both starting reactants and final products, and must predict the intermediate steps.}
\small
\label{tab:3}
\begin{tabular}{lccc}
\toprule
 & \textbf{A} & \textbf{B} & \textbf{C} \\
\midrule
GPT-5           & 21.8\% & 4.6\%  & 0.0\% \\
Gemini-3.0-Pro  & 44.9\% & 20.6\% & 4.5\% \\
Qwen3-235B-A22B & 7.7\%  & 3.1\%  & 0.0\% \\
Qwen3-30B-A3B   & 1.3\%  & 0.0\%  & 0.0\% \\
\midrule
ChemDFM-R-14B \citep{zhao2025chemdfmr}  & 1.3\%  & 0.0\%  & 0.0\% \\
\midrule
Ours            & \textbf{10.3\%} & 1.5\% & 0.0\% \\
\bottomrule
\end{tabular}
\end{table}

In contrast, our model achieves 8.3\% Top-1 on the pathway \& product task (Tab.~\ref{tab:2}), surpassing both the base \texttt{Qwen3-30B-A3B} (6.9\%) and the specialized FlowER model (5.1\%). Our Top-8 accuracy of 15.4\% demonstrates strong predictive diversity---the model can explore multiple plausible pathways, a desirable property for mechanism hypothesis generation. On the pathway-only task (Tab.~\ref{tab:3}), where the model receives both endpoints and must predict intermediates, our model reaches 10.3\% Top-1 on Set~A, surpassing \texttt{Qwen3-235B-A22B} (7.7\%) and dramatically outperforming the base \texttt{Qwen3-30B} (1.3\%). 

On Set~B (graduate-level problems), all models fall below 5\% on both tasks, indicating that this difficulty tier approaches the noise floor for current 30B-scale models. Set~C (research-level) remains largely unsolved by all models, including \texttt{Gemini-3.0-Pro} which achieves only 1.0\% on pathway \& product and 4.5\% on pathway-only. These results suggest that Sets~B and C require fundamentally broader reaction type coverage in training data, particularly pericyclic and radical mechanisms that are underrepresented in current datasets. Complete results for the step-wise only prediction task are provided in Appendix~\ref{app:results}.

\subsection{Case Studies}

\begin{figure*}
    \centering
    \includegraphics[width=1\linewidth]{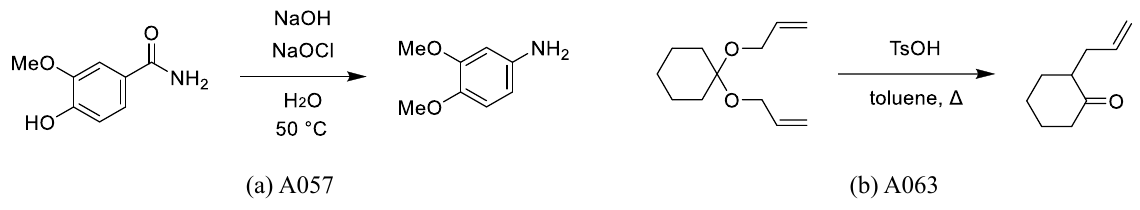}
    \caption{Representative reaction mechanism prediction for FukuyamaBench cases.}
    \label{fig:4}
\end{figure*}

To qualitatively evaluate our model's reasoning capabilities, we analyze two representative multi-step transformations from FukuyamaBench (Figure~\ref{fig:4}). Case A057 involves a Hofmann-style rearrangement. Our model successfully navigates the complete mechanistic backbone—from $N$-chlorination to the critical isocyanate intermediate—whereas GPT-5 becomes derailed by overly explicit ionic bookkeeping, and Qwen3-30B-A3B fundamentally misidentifies the reaction as direct amide hydrolysis. Similarly, in Case A063, our model correctly utilizes acid protonation as an activation step to drive a productive $\beta$-elimination, yielding the target conjugated enol-ether. Conversely, GPT-5 prematurely halts at oxonium formation, treating activation as the terminal state, while Qwen3 hallucinates incorrect downstream carbocation pathways. As detailed further in the Appendix~\ref{app:case}, these cases demonstrate that our mechanism-aware training enables the model to reliably track complex chemical trajectories, successfully avoiding the superficial dead-ends and mechanistic hallucinations that plague baseline models. 

\section{Conclusion}

We presented a framework for teaching LLMs to reason about organic reaction mechanisms at the elementary-step level. Our contributions are threefold: (1)~\textbf{FukuyamaBench}, the first mechanism reasoning benchmark with atom mapping, conservation validation, and multi-task evaluation across 319 problems spanning undergraduate to research difficulty, where all extracted mechanisms and checkpoints are expert-validated; (2)~a \textbf{distribution-matched reasoning dataset} constructed from FlowER, oMe-Silver, PMech, and expert-curated sources, with structured forward-predictive annotations, validated SMILES, atom-ID audits, and prompt templates to be released for transparency; and (3)~\textbf{empirical validation} that mechanism-aware post-training on \texttt{Qwen3-30B-A3B} achieves 8.3\% exact pathway match on Set~A, surpassing the base model (6.9\%) and the specialized FlowER model (5.1\%), with a Top-8 accuracy of 15.4\% demonstrating strong predictive diversity.

\textbf{Limitations.} A significant gap remains to frontier models such as \texttt{Gemini-3.0-Pro} (22.3\%), and Sets~B and C remain largely unsolved by all models, suggesting that complex rearrangements, pericyclic, and radical mechanisms require both richer training data and multi-modal capability to address. In addition, exact-match evaluation may under-credit alternative but chemically plausible pathways \citep{zagribelnyy2026singleanswerenoughrethinking}.

\textbf{Future directions} include scaling training data with broader reaction type coverage, incorporating multi-modal inputs (e.g., molecular graph), extending the framework to retrosynthetic pathway planning, and developing agentic systems that construct intermediate reaction graphs with physically grounded, ML-potential-based feedback for pathway selection and reinforcement learning.

\section{Supporting Information}

Additional experimental details. (PDF)

\section{Acknowledgment}
This work was supported by funds from the National Institutes of Health (R01GM147673, R01GM149705) and the National Science Foundation (1955260). The authors would like to thank the computing resources provided by the Center for Research Computing (facility RRID: SCR\_022735) at the University of Pittsburgh (NSF award number OAC-2117681), and the Pittsburgh Supercomputer Center (grant number BIO210185). This work was also supported in part by the University of Florida (UF Startup Fund and UF-NVIDIA Artificial Intelligence and Complex Computational Research Award to Y. L.), National Institutes of Health (R21EB037868 to Y. L.). We acknowledge UFIT Research Computing for providing computational resources.

\section{Data and Software Availability}
All training code, configurations, and evaluation scripts are available at \url{https://github.com/HaCTang/ReactionMechanismReasoning}. The reported results correspond to commit \texttt{7c1cb69}. The released training dataset are available on Hugging Face at \url{https://huggingface.co/datasets/Haocheng1/mech-infer-train-v1}. The code and dataset is released under the Apache License 2.0; other datasets, base models, and fine-tuned weights follow the licenses or availability conditions of their original sources.

\section{AUTHOR INFORMATION }

\subsection{Corresponding Author} 

Junmei Wang - Department of Pharmaceutical Sciences and Computational Chemical Genomics Screening Center, School of Pharmacy, University of Pittsburgh, Pittsburgh, PA, 15261, USA. Email: juw79@pitt.edu 

Yanjun Li - Department of Medicinal Chemistry, Center for Natural Products, Drug Discovery and Development, Department of Computer and Information Science and Engineering, University of Florida, Gainesville, FL, 15261, USA. Email: yanjun.li@ufl.edu

\subsection{Other Authors} 
Xingyu Dang - Department of Computer Science, Princeton University, Princeton, NJ, 08540, USA. Email: xingyu.dang@princeton.edu

Haocheng Tang - Department of Pharmaceutical Sciences and Computational Chemical Genomics Screening Center, School of Pharmacy, University of Pittsburgh, Pittsburgh, PA, 15261, USA; Khoury College of Computer Science, Northeastern University, Boston, MA, 02115, USA. Email: tang.haoc@northeastern.edu

\subsection{Author Contributions} 
H.T. and X.D. designed the research, performed the research and analyzed the data and wrote the manuscript. J.W. and Y.L. provided computing resources. All authors reviewed the manuscript.

\subsection{Conflict of Interest}
The authors declare no competing financial interests.

\let\thefootnote\relax\footnotetext{*Corresponding author: Junmei Wang (\texttt{JUW79@pitt.edu})}
    
\bibliography{example_paper}

\appendix
\clearpage 
\subsection{Performance on FukuyamaBench}

\begin{table}[!htbp]
\centering
\caption{The result of step-wise only prediction task}
\resizebox{\linewidth}{!}{
\begin{tabular}{lccccccccc}
\toprule
 & \multicolumn{3}{c}{\textbf{A}} & \multicolumn{3}{c}{\textbf{B}} & \multicolumn{3}{c}{\textbf{C}} \\
\cmidrule(lr){2-4} \cmidrule(lr){5-7} \cmidrule(lr){8-10}
 & Top-1 & Top-3 & Top-5 & Top-1 & Top-3 & Top-5 & Top-1 & Top-3 & Top-5 \\
\midrule
FlowER \citep{joung2025flower}         & 19.7\% & 30.1\% & 32.9\% & 13.2\% & 22.5\% & 24.2\% & 10.7\% & 18.0\% & 20.1\% \\
GPT-5           & 34.5\% &   -    &   -    & 24.3\% &   -    &   -    & 17.7\% &   -    &  -     \\
Gemini-3.0-Pro  & 49.8\% &   -    &   -    & 42.9\% &   -    &   -    & 32.7\% &   -    &  -     \\
Qwen3-235B-A22B & 15.5\% &   -    &   -    & 9.2\%  &   -    &   -    & 7.3\%  &   -    &  -     \\
Qwen3-30B-A3B   & 1.7\%  &   -    &   -    & 2.9\%  &   -    &   -    & 2.2\%  &   -    &  -     \\
\midrule
ChemDFM-R-14B \citep{zhao2025chemdfmr}  & 1.6\%  &  1.6\% &  1.6\% & 0.0\%  & 0.0\%  &  0.0\% & 0.8\%  &  0.8\% &  0.8\%  \\
\midrule
Ours   & 8.9\%  & 13.6\% & 16.4\% & 4.8\%  & 8.1\%  & 10.1\% & 5.3\%  & 7.9\%  & 9.2\%  \\
\bottomrule
\end{tabular}
}
\end{table}

The step-wise prediction task evaluates whether the model can predict the next elementary step given the current intermediate state. FlowER achieves 19.7\% Top-1 on Set~A, outperforming our model (8.9\%). This is expected: FlowER is a specialized flow-matching model trained explicitly for single-step prediction on the FlowER reaction space, whereas our model is trained for full-pathway generation and is evaluated here on a task format it was not optimized for. Nevertheless, our model substantially outperforms all other fine-tuned LLMs on this task, including \texttt{ChemDFM-R-14B} (1.6\%) and the base \texttt{Qwen3-30B-A3B} (1.7\%), demonstrating that mechanism-aware training transfers across evaluation granularities.

\subsection{Performance of Gemini on FukuyamaBench based on reaction types}

We further analyze Gemini-3.0-Pro at the elementary-step level across reaction classes. The reaction-type distribution shows that FukuyamaBench covers diverse mechanisms, with proton transfer, addition, elimination, substitution, and pericyclic steps forming the major categories, while radical and electron-transfer steps are relatively sparse. Gemini exhibits strong class-dependent variation. In the pathway \& product task, coordination, electron transfer, addition, and proton transfer achieve relatively higher Top-1 recall, whereas elimination and rearrangement are much more challenging. Step-wise evaluation substantially improves recall for most classes, indicating that many errors arise from long-horizon trajectory construction rather than local step recognition alone. Our mechanism-trained model follows a similar trend but improves over the base \texttt{Qwen3-30B-A3B} on mechanism-sensitive categories, especially in full-pathway generation, suggesting that mechanism-aware post-training helps the model better track reactive sites, intermediate evolution, and chemically valid step transitions, although challenging classes such as rearrangement, pericyclic reactions, and long elimination sequences remain difficult.

\begin{table}[ht]
\centering
\caption{Gemini-3.0-Pro elementary-step per-class performance across task modes.}
\resizebox{0.85\linewidth}{!}{
\begin{tabular}{llrrrrr}
\toprule
Mode & Reaction class & Units & Top-1 hit & Top-1 recall & Top-8 hit & Top-8 recall \\
\midrule
\multirow{10}{*}{Pathway \& product} & Proton Transfer & 507 & 78 & 15.4\% & 130 & 25.6\% \\
 & Addition & 423 & 73 & 17.3\% & 100 & 23.6\% \\
 & Elimination & 325 & 5 & 1.5\% & 14 & 4.3\% \\
 & Substitution & 272 & 26 & 9.6\% & 41 & 15.1\% \\
 & Pericyclic & 195 & 21 & 10.8\% & 28 & 14.4\% \\
 & Cleavage & 98 & 9 & 9.2\% & 16 & 16.3\% \\
 & Rearrangement & 79 & 1 & 1.3\% & 4 & 5.1\% \\
 & Coordination & 52 & 12 & 23.1\% & 17 & 32.7\% \\
 & Electron Transfer & 24 & 5 & 20.8\% & 7 & 29.2\% \\
 & Radical & 22 & 3 & 13.6\% & 5 & 22.7\% \\
\midrule
\multirow{10}{*}{Pathway only} & Proton Transfer & 507 & 64 & 12.6\% &  &  \\
 & Addition & 423 & 81 & 19.1\% &  &  \\
 & Elimination & 325 & 16 & 4.9\% &  &  \\
 & Substitution & 272 & 24 & 8.8\% &  &  \\
 & Pericyclic & 195 & 29 & 14.9\% &  &  \\
 & Cleavage & 98 & 4 & 4.1\% &  &  \\
 & Rearrangement & 79 & 9 & 11.4\% &  &  \\
 & Coordination & 52 & 10 & 19.2\% &  &  \\
 & Electron Transfer & 24 & 0 & 0.0\% &  &  \\
 & Radical & 22 & 3 & 13.6\% &  &  \\
\midrule
\multirow{10}{*}{Step-wise} & Proton Transfer & 507 & 269 & 53.1\% &  &  \\
 & Addition & 423 & 217 & 51.3\% &  &  \\
 & Elimination & 325 & 115 & 35.4\% &  &  \\
 & Substitution & 272 & 119 & 43.8\% &  &  \\
 & Pericyclic & 195 & 75 & 38.5\% &  &  \\
 & Cleavage & 98 & 29 & 29.6\% &  &  \\
 & Rearrangement & 79 & 24 & 30.4\% &  &  \\
 & Coordination & 52 & 28 & 53.8\% &  &  \\
 & Electron Transfer & 24 & 3 & 12.5\% &  &  \\
 & Radical & 22 & 10 & 45.5\% &  &  \\
\bottomrule
\end{tabular}
}
\end{table}

\begin{table}[ht]
\centering
\caption{Our model's checkpoint-level Top-8 recall by reaction class across FukuyamaBench sets on pathway \& product prediction task.}
\resizebox{\linewidth}{!}{
\begin{tabular}{lrrrrrr}
\toprule
Reaction class & Set A ckpts & Set A Top-8 recall & Set B ckpts & Set B Top-8 recall & Set C ckpts & Set C Top-8 recall \\
\midrule
Addition & 91 & 27.5\% & 142 & 7.0\% & 148 & 6.8\% \\
Cleavage & 10 & 50.0\% & 33 & 6.1\% & 34 & 2.9\% \\
Coordination & 8 & 0.0\% & 13 & 15.4\% & 12 & 16.7\% \\
Electron Transfer & 2 & 0.0\% & 13 & 23.1\% & 3 & 0.0\% \\
Elimination & 50 & 10.0\% & 79 & 3.8\% & 58 & 6.9\% \\
Pericyclic & 14 & 0.0\% & 36 & 0.0\% & 89 & 0.0\% \\
Proton Transfer & 30 & 60.0\% & 31 & 29.0\% & 24 & 20.8\% \\
Radical & 2 & 50.0\% & 9 & 0.0\% & 6 & 16.7\% \\
Rearrangement & 6 & 16.7\% & 18 & 0.0\% & 33 & 0.0\% \\
Substitution & 39 & 30.8\% & 79 & 13.9\% & 86 & 7.0\% \\
\bottomrule
\end{tabular}
}
\end{table}

\begin{figure}[ht]
    \centering
    \includegraphics[width=0.7\linewidth]{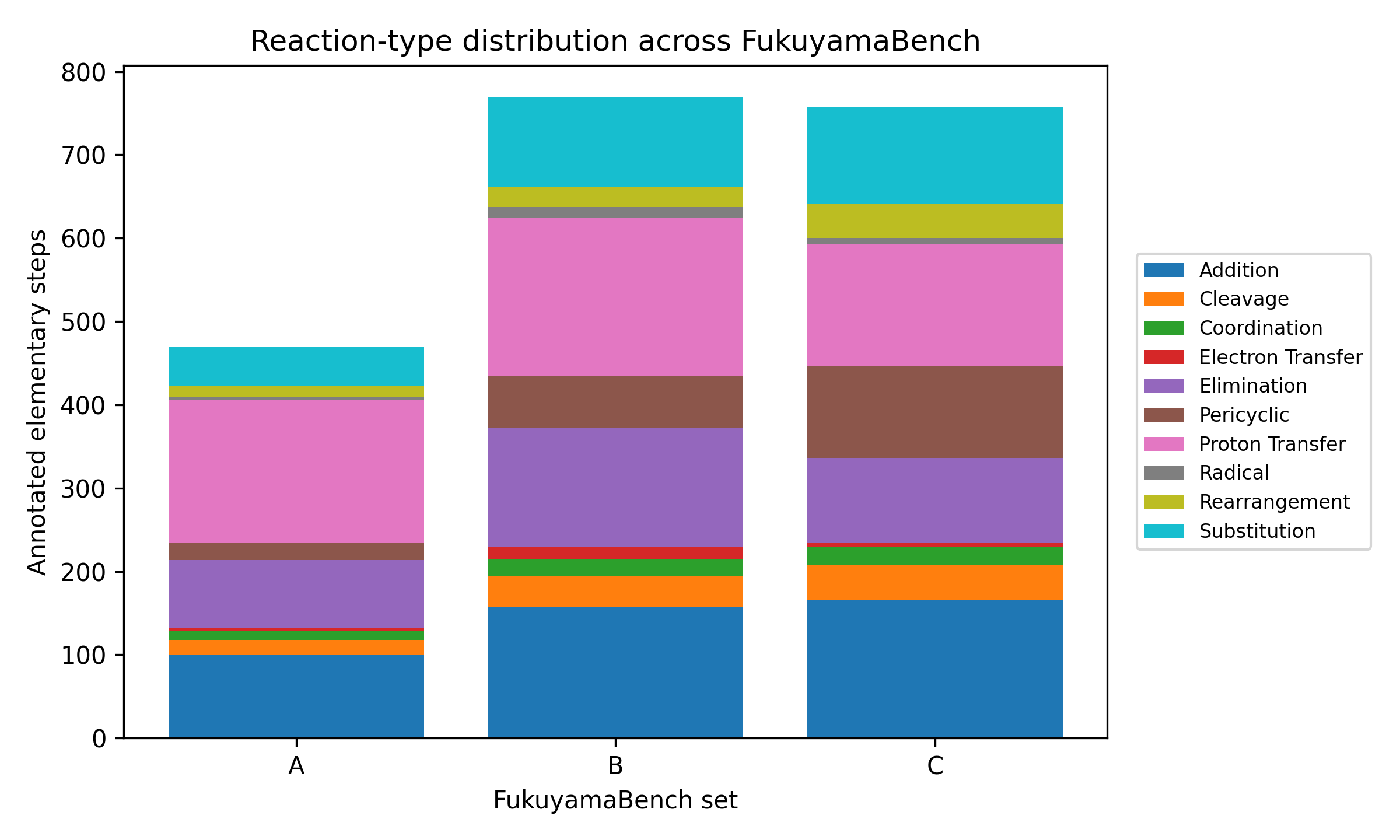}
    \caption{Full reaction mechanism of FukuyamaBench A063.}
    \label{fig:s4}
\end{figure}

\clearpage
\newpage

\section{Dataset Processing Details}

\subsection{FukuyamaBenchmark}

FukuyamaBench was constructed from the \textit{Advanced Organic Reaction Mechanism} workbook, which provides paired reaction questions and reference mechanism solutions. The original materials contain reaction schemes and hand-drawn mechanism figures rather than directly machine-readable trajectories. We therefore first decomposed the source materials into reaction images and mechanism images. For each reaction image, we extracted the reactants, products, and reaction conditions into textual molecular representations. The extracted reaction texts were then manually checked and corrected by chemistry domain experts to ensure consistency with the original problem.

For the mechanism side, Gemini was used only as an assistant for extracting and formatting information from the provided mechanism figures, rather than as an unconstrained generator of benchmark ground truth. In this annotation setting, Gemini was given substantially richer information than the models evaluated on the benchmark, including the reactant and product SMILES/SMARTS, reaction conditions, and the full reference mechanism image. Based on each question as a boundary, the parsing accuracy for Set A, B, and C was 63/78, 76/131, and 34/110 respectively. Each elementary reaction step was further inspected by RDKit to ensure step-wise atomic and electronic conservation as shown in Example \ref{example:1}. The resulting mechanism trajectories were reviewed and corrected by domain experts. Subsequently, using the generated mechanism trajectories as inputs,intermediate checkpoints for the pathway \& product prediction task and the  pathway-only prediction task were derived according to predefined coarse-grained rules (See Section \ref{para:ckpt}). Checkpoints were also verified by experts.

\begin{figure*}[ht]
    \centering
    \includegraphics[width=0.7\linewidth]{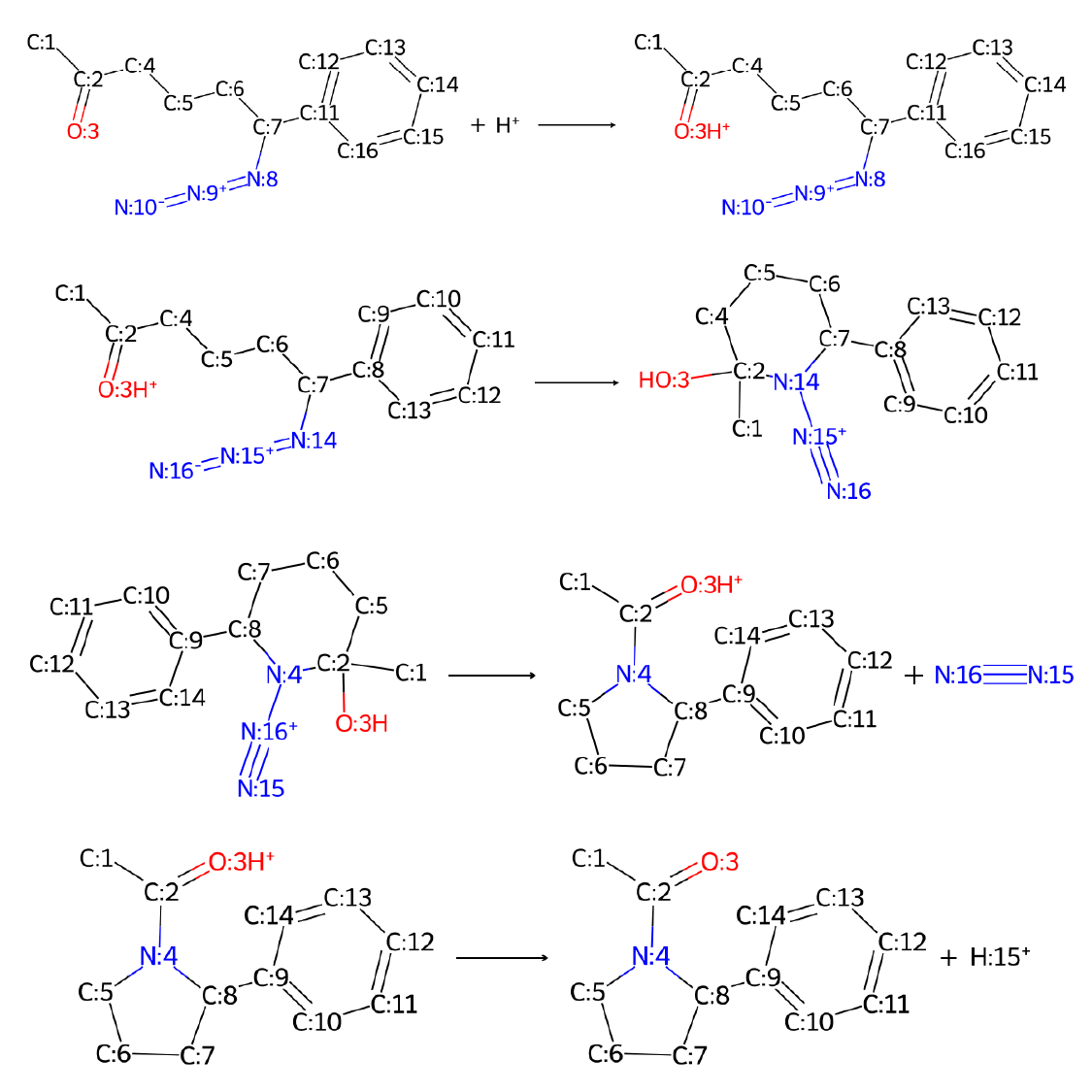}
    \caption{Visualization of mapped RXN SMILES in B111-a.}
    \label{fig:s5}
\end{figure*}

Finally, atom mapping was assigned using RxnMapper and verified by experts, so that backbone atom correspondence and reactive sites could be traced across elementary steps as shown in Example \ref{example:2} and fig.~\ref{fig:s5}. This process yielded atom-mapped trajectories with explicit intermediates, products, reaction conditions, and checkpoint-level mechanistic states. The benchmark was then organized into three tasks: step-wise only prediction, pathway-only prediction, and pathway \& product prediction. Importantly, during benchmark evaluation, models are given only the task-specific inputs such as reactants/intermediate states and reaction conditions; they are not given the product SMILES, SMARTS, or reference mechanism images used during dataset construction.

\renewcommand{\lstlistingname}{Example}
\begin{lstlisting}[
  frame=single,
  caption={B111-a mechanism trajectory in FukuyamaBench.},
  label={example:1},
  captionpos=t,
  basicstyle=\ttfamily\scriptsize,
  lineskip=-1pt,
  breaklines=true,
  showstringspaces=false,
  literate={"}{{\char34}}1 {°}{{$^\circ$}}1
]
{
    "reactions": [
        {
            "reaction_id": "1",
            "reactants": [
                {
                    "smiles": "CC(=O)CCCC(N=[N+]=[N-])c1ccccc1",
                    "label": "None",
                    "electron_number": 84,
                    "charge": 0
                }
            ],
            "conditions": [
                {
                    "role": "reagent",
                    "text": "TfOH",
                    "reaction_step": "1"
                },
                {
                    "role": "solvent",
                    "text": "CH2Cl2",
                    "reaction_step": "1"
                },
                {
                    "role": "temperature",
                    "text": "0 °C",
                    "reaction_step": "1"
                }
            ],
            "products": [
                {
                    "smiles": "CC(=O)N1CCCC1c1ccccc1",
                    "label": "None",
                    "electron_number": 74,
                    "charge": 0
                }
            ]
        }
    ],
    "mechanism": [
        {
            "mechanism_id": "1",
            "reactants": [
                {
                    "smiles": "CC(=O)CCCC(N=[N+]=[N-])c1ccccc1",
                    "label": "substrate",
                    "electron_number": 84,
                    "charge": 0
                },
                {
                    "smiles": "[H+]",
                    "label": "reagent",
                    "electron_number": 0,
                    "charge": 1
                }
            ],
            "products": [
                {
                    "smiles": "CC(=[OH+])CCCC(N=[N+]=[N-])c1ccccc1",
                    "label": "intermediate",
                    "electron_number": 84,
                    "charge": 1
                }
            ]
        },
        {
            "mechanism_id": "2",
            "reactants": [
                {
                    "smiles": "CC(=[OH+])CCCC(N=[N+]=[N-])c1ccccc1",
                    "label": "intermediate",
                    "electron_number": 84,
                    "charge": 1
                }
            ],
            "products": [
                {
                    "smiles": "N#[N+]N1(C(C)(O)CCCC1c1ccccc1)",
                    "label": "intermediate",
                    "electron_number": 84,
                    "charge": 1
                }
            ]
        },
        {
            "mechanism_id": "3",
            "reactants": [
                {
                    "smiles": "N#[N+]N1(C(C)(O)CCCC1c1ccccc1)",
                    "label": "intermediate",
                    "electron_number": 84,
                    "charge": 1
                }
            ],
            "products": [
                {
                    "smiles": "CC(=[OH+])N1CCCC1c1ccccc1",
                    "label": "intermediate",
                    "electron_number": 74,
                    "charge": 1
                },
                {
                    "smiles": "N#N",
                    "label": "leaving",
                    "electron_number": 10,
                    "charge": 0
                }
            ]
        },
        {
            "mechanism_id": "4",
            "reactants": [
                {
                    "smiles": "CC(=[OH+])N1CCCC1c1ccccc1",
                    "label": "intermediate",
                    "electron_number": 74,
                    "charge": 1
                }
            ],
            "products": [
                {
                    "smiles": "CC(=O)N1CCCC1c1ccccc1",
                    "label": "product",
                    "electron_number": 74,
                    "charge": 0
                },
                {
                    "smiles": "[H+]",
                    "label": "leaving",
                    "electron_number": 0,
                    "charge": 1
                }
            ]
        }
    ]
}

\end{lstlisting}

\begin{lstlisting}[
  frame=single,
  caption={Mapped RXN SMILES of B111-a in FukuyamaBench.},
  label={example:2},
  captionpos=t,
  basicstyle=\ttfamily\scriptsize,
  lineskip=-1pt,
  breaklines=true,
  showstringspaces=false,
  literate={"}{{\char34}}1 {°}{{$^\circ$}}1
]
{
    "mechanism_rxn": [
        {
            "step": "1",
            "rxn_smiles": "CC(=O)CCCC(N=[N+]=[N-])c1ccccc1.[H+]>>CC(=[OH+])CCCC(N=[N+]=[N-])c1ccccc1",
            "mapped_rxn": "[CH3:1][C:2](=[O:3])[CH2:4][CH2:5][CH2:6][CH:7]([N:8]=[N+:9]=[N-:10])[c:11]1[cH:12][cH:13][cH:14][cH:15][cH:16]1.[H+]>>[CH3:1][C:2](=[OH+:3])[CH2:4][CH2:5][CH2:6][CH:7]([N:8]=[N+:9]=[N-:10])[c:11]1[cH:12][cH:13][cH:14][cH:15][cH:16]1"
        },
        {
            "step": "2",
            "rxn_smiles": "CC(=[OH+])CCCC(N=[N+]=[N-])c1ccccc1>>N#[N+]N1(C(C)(O)CCCC1c1ccccc1)",
            "mapped_rxn": "[CH3:1][C:2](=[OH+:3])[CH2:4][CH2:5][CH2:6][CH:7]([c:8]1[cH:9][cH:10][cH:11][cH:12][cH:13]1)[N:14]=[N+:15]=[N-:16]>>[CH3:1][C:2]1([OH:3])[CH2:4][CH2:5][CH2:6][CH:7]([c:8]2[cH:9][cH:10][cH:11][cH:12][cH:13]2)[N:14]1[N+:15]#[N:16]"
        },
        {
            "step": "3",
            "rxn_smiles": "N#[N+]N1(C(C)(O)CCCC1c1ccccc1)>>CC(=[OH+])N1CCCC1c1ccccc1.N#N",
            "mapped_rxn": "[CH3:1][C:2]1([OH:3])[N:4]([N+:16]#[N:15])[CH:8]([c:9]2[cH:10][cH:11][cH:12][cH:13][cH:14]2)[CH2:7][CH2:6][CH2:5]1>>[CH3:1][C:2](=[OH+:3])[N:4]1[CH2:5][CH2:6][CH2:7][CH:8]1[c:9]1[cH:10][cH:11][cH:12][cH:13][cH:14]1.[N:15]#[N:16]"
        },
        {
            "step": "4",
            "rxn_smiles": "CC(=[OH+])N1CCCC1c1ccccc1>>CC(=O)N1CCCC1c1ccccc1.[H+]",
            "mapped_rxn": "[CH3:1][C:2](=[OH+:3])[N:4]1[CH2:5][CH2:6][CH2:7][CH:8]1[c:9]1[cH:10][cH:11][cH:12][cH:13][cH:14]1>>[CH3:1][C:2](=[O:3])[N:4]1[CH2:5][CH2:6][CH2:7][CH:8]1[c:9]1[cH:10][cH:11][cH:12][cH:13][cH:14]1.[H+:15]"
        }
    ]
}

\end{lstlisting}

\subsection{Training Dataset}

For data construction, we first transformed raw reaction strings to a reaction-wise trajectory representation. Each reaction is stored as an ordered list of elementary steps with explicit step index, reactants, products, and raw reaction string. This preserves causal order for mechanism supervision. Identity transitions are removed during formatting, so only chemically active steps are retained. We then perform pathway-level reasoning annotation, where each sample starts from initial reactants and predicts the complete multi-step trajectory. The annotator is constrained to produce both free-form forward reasoning and a structured result JSON. To improve robustness at scale, we use bounded asynchronous generation, retry with exponential backoff, periodic checkpointing, and automatic resume; invalid entries, such as empty trajectories or abnormal reaction identifiers, are filtered before annotation. The valid annotations are finally converted into a standard multi-turn instruction format for supervised fine-tuning. 

\clearpage
\newpage

\section{Training Details}
\label{app:training}

All training is conducted on 8$\times$ NVIDIA B200 GPUs (180\,GB each) using PyTorch 2.9 with BF16 mixed precision and FlashAttention-2.

\begin{table}[!htbp]
\centering
\caption{SFT hyperparameters}
\begin{tabular}{ll}
\toprule
\textbf{Parameter} & \textbf{Value} \\
\midrule
Base model & Qwen3-30B-A3B-Instruct \\
Training data & 3,650 HQ samples (2,353 FlowER + 1,297 external) \\
Precision & BF16\\
Optimizer & AdamW \\
Learning rate & $1 \times 10^{-5}$ (cosine decay) \\
Epochs & 3 (687 steps) \\
Max sequence length & 16,384 tokens \\
Batch size & 8 \\
Wall time & $\sim$2 hours \\
\bottomrule
\end{tabular}
\end{table}

\begin{table}[!htbp]
\centering
\caption{RL hyperparameters}
\begin{tabular}{ll}
\toprule
\textbf{Parameter} & \textbf{Value} \\
\midrule
Base model & SFT checkpoint above \\
RL framework & verl \\
Algorithm & GSPO  \\
Training data & 2,000 samples (1,098 external + 902 FlowER) \\
Group size ($n$) & 8 \\
Batch size & 16 \\
Learning rate & $1 \times 10^{-6}$ \\
Weight decay & 0.1 \\
Clip ratio & $\epsilon_\text{low} = 3 \times 10^{-4}$, $\epsilon_\text{high} = 4 \times 10^{-4}$ \\
KL penalty & None (tight clipping suffices) \\
Max response length & 12,288 tokens \\
Epochs & 3 (369 steps) \\
Wall time & $\sim$20 hours \\
\bottomrule
\end{tabular}
\end{table}

\clearpage
\newpage
\section{Prompts}

\paragraph{The prompt of step-wise only prediction}:

\begin{lstlisting}[frame=single, breaklines=true, basicstyle=\ttfamily\small, literate={"}{{\char34}}1]

student_system_prompt: |
  Role: You are a Chemical Mechanism Reasoning Engine. Task: Given the current chemical state (SMILES with atom mapping), predict the next elementary transition.
  
  Output Protocol:
    1. Provide your step-by-step reasoning under the header ## Reasoning.
    2. Provide the final prediction and metadata in a single JSON block under the header ## Result.
    
    JSON Specification: The JSON must contain exactly these keys:
    "next_state": The SMILES string of the next intermediate or product.
    "reaction_class": The category of the reaction (e.g., "SN2", "Proton Transfer").
    "is_terminal": Boolean, true if this is the stable final product of the trajectory.
    "reactive_sites": A dictionary with "nucleophile_id" and "electrophile_id" keys, means the nucleophile and electrophile atoms that are involved in the reaction.

student_template_prompt: |
  Reactants:   
  COC(=O)c1cccc([N+](=O)[O-])c1.[OH-]
  [CH3:1][O:2][C:3](=[O:5])[c:6]1[cH:7][cH:8][cH:9][c:10]([N+:11](=[O:12])[O-:13])[cH:14]1.[OH-:4]

\end{lstlisting}

\paragraph{The prompt of pathway-only prediction}:

\begin{lstlisting}[frame=single, breaklines=true, basicstyle=\ttfamily\small, literate={"}{{\char34}}1]
student_system_prompt: |
  Role: You are an expert Computational Chemist and Mechanism Designer. You prioritize chemical validity and strict SMILES syntax.

  Task: Given the starting reactants, products (SMILES with atom mapping) and reaction conditions, predict the complete multi-step reaction mechanism.

  ## Instructions

  1. **Reasoning Section (Mandatory)**
     Analyze the mechanism step-by-step. For each step, you MUST follow this structured process:

     A. **Mechanism Analysis**: 
        - Analyze electron flow (Source -> Sink).
        - Identify bond formation/breaking using Atom Map IDs.

     B. **Atom State Definition (CRITICAL)**:
        - Before generating SMILES, explicitly define the state of the reaction centers.
        - Format: "Atom [ID]: [Element], [Valence/Bonds], [Charge]"
        - *Example*: "Atom :5 (Oxygen): 1 bond, 3 lone pairs -> Charge -1."
        - *Example*: "Atom :3 (Carbon): 4 single bonds -> Neutral."

     C. **SMILES Syntax Translation**:
        - **Rule Check**: If Atom X is defined as Negative, the SMILES segment MUST be `[Element-:ID]`. 
        - **Sanity Check**: Does the draft SMILES imply a pentavalent carbon (5 bonds) or a negative carbon with 4 bonds? If yes, CORRECT IT immediately by moving the charge to the heteroatom.

     D. **Predicted Intermediate**:
        - The final SMILES string derived from the logic above.

  2. **Result Section**
     Provide a JSON array with each elementary step. Continue until you reach the terminal
     stable product.

  ## Output Format

  ## Reasoning

  **Step 1:**
  [Your analysis of the first elementary step...]
  Predicted intermediate: [describe or show SMILES]

  **Step 2:**
  [Your analysis...]
  ...

  **Step N (Terminal):**
  [Analysis of the final step leading to stable product...]

  ## Result

  ```json
  [
    {
      "step_id": 1,
      "reaction_type": "e.g., Nucleophilic Addition",
      "description": "Brief description of what happens",
      "reactive_sites": {"nucleophile": "[O:5]", "electrophile": "[C:3]"},
      "product_smiles": "SMILES of the intermediate/product",
      "is_terminal": false
    },
    ...
    {
      "step_id": N,
      "reaction_type": "...",
      "description": "...",
      "reactive_sites": {...},
      "product_smiles": "SMILES of final product",
      "is_terminal": true
    }
  ]
  ```

student_template_prompt: |
  Starting Reactants:
  COC(=O)c1cccc([N+](=O)[O-])c1.[OH-]
  [CH3:1][O:2][C:3](=[O:5])[c:6]1[cH:7][cH:8][cH:9][c:10]([N+:11](=[O:12])[O-:13])[cH:14]1.[OH-:4]

  Ending Products:
  O=C(O)c1cccc([N+](=O)[O-])c1
  [O:1]=[C:2]([OH:3])[c:4]1[cH:5][cH:6][cH:7][c:8]([N+:9](=[O:10])[O-:11])[cH:12]1

  Reaction Conditions:
  reagent: NaOH (step 1)
  solvent: H2O (step 1)
  temperature: reflux (step 1)
  reagent: aq HCl (step 2)

  **CRITICAL OUTPUT GUIDELINES:**
  
  **1. Avoid Pentavalent Carbons / Invalid Charges:**
  - Common Mistake: Placing a negative charge on a Carbon that already has 4 bonds (e.g., `[C-](O)(O)`). This violates the Octet Rule.
  - Correction: The charge usually belongs to the adjacent Oxygen.
  
  **2. One-Shot Example (Learn from this):**
  - **Context**: Hydroxide attacks a Carbonyl.
  - **Reasoning**: Electrons move to Oxygen.
  - **INCORRECT SMILES**: `...[C-](O)(O)...` (WRONG: Carbon has 5 valences/charge errors)
  - **CORRECT SMILES**: `...C([O-])(O)...` (RIGHT: Charge on Oxygen)

  Predict the complete reaction mechanism step-by-step.
\end{lstlisting}

\paragraph{The prompt of pathway \& product prediction}:
\begin{lstlisting}[frame=single, breaklines=true, basicstyle=\ttfamily\small, literate={"}{{\char34}}1]
student_system_prompt: |
    Role: You are an expert Computational Chemist and Mechanism Designer. You prioritize chemical validity and strict SMILES syntax.
    
    Task: Given the starting reactants (SMILES with atom mapping) and reaction conditions, predict the complete multi-step reaction mechanism.
    
    ## Instructions
    
    1. **Reasoning Section (Mandatory)**
       Analyze the mechanism step-by-step. For each step, you MUST follow this structured process:
    
       A. **Mechanism Analysis**: 
          - Analyze electron flow (Source -> Sink).
          - Identify bond formation/breaking using Atom Map IDs.
    
       B. **Atom State Definition (CRITICAL)**:
          - Before generating SMILES, explicitly define the state of the reaction centers.
          - Format: "Atom [ID]: [Element], [Valence/Bonds], [Charge]"
          - *Example*: "Atom :5 (Oxygen): 1 bond, 3 lone pairs -> Charge -1."
          - *Example*: "Atom :3 (Carbon): 4 single bonds -> Neutral."
    
       C. **SMILES Syntax Translation**:
          - **Rule Check**: If Atom X is defined as Negative, the SMILES segment MUST be `[Element-:ID]`. 
          - **Sanity Check**: Does the draft SMILES imply a pentavalent carbon (5 bonds) or a negative carbon with 4 bonds? If yes, CORRECT IT immediately by moving the charge to the heteroatom.
    
       D. **Predicted Intermediate**:
          - The final SMILES string derived from the logic above.
    
    2. **Result Section**
       Provide a JSON array with each elementary step. Continue until you reach the terminal
       stable product.
    
    ## Output Format
    
    ## Reasoning
    
    **Step 1:**
    [Your analysis of the first elementary step...]
    Predicted intermediate: [describe or show SMILES]
    
    **Step 2:**
    [Your analysis...]
    ...
    
    **Step N (Terminal):**
    [Analysis of the final step leading to stable product...]
    
    ## Result
    
    ```json
    [
      {
        "step_id": 1,
        "reaction_type": "e.g., Nucleophilic Addition",
        "description": "Brief description of what happens",
        "reactive_sites": {"nucleophile": "[O:5]", "electrophile": "[C:3]"},
        "product_smiles": "SMILES of the intermediate/product",
        "is_terminal": false
      },
      ...
      {
        "step_id": N,
        "reaction_type": "...",
        "description": "...",
        "reactive_sites": {...},
        "product_smiles": "SMILES of final product",
        "is_terminal": true
      }
    ]
    ```

student_template_prompt: |
  Starting Reactants:
  COC(=O)c1cccc([N+](=O)[O-])c1.[OH-]
  [CH3:1][O:2][C:3](=[O:5])[c:6]1[cH:7][cH:8][cH:9][c:10]([N+:11](=[O:12])[O-:13])[cH:14]1.[OH-:4]

  Reaction Conditions:
  reagent: NaOH (step 1)
  solvent: H2O (step 1)
  temperature: reflux (step 1)
  reagent: aq HCl (step 2)

  **CRITICAL OUTPUT GUIDELINES:**
  
  **1. Avoid Pentavalent Carbons / Invalid Charges:**
  - Common Mistake: Placing a negative charge on a Carbon that already has 4 bonds (e.g., `[C-](O)(O)`). This violates the Octet Rule.
  - Correction: The charge usually belongs to the adjacent Oxygen.
  
  **2. One-Shot Example (Learn from this):**
  - **Context**: Hydroxide attacks a Carbonyl.
  - **Reasoning**: Electrons move to Oxygen.
  - **INCORRECT SMILES**: `...[C-](O)(O)...` (WRONG: Carbon has 5 valences/charge errors)
  - **CORRECT SMILES**: `...C([O-])(O)...` (RIGHT: Charge on Oxygen)

  Predict the complete reaction mechanism step-by-step.

\end{lstlisting}

\paragraph{The prompt of Gemini extraction of reactions in FukuyamaBench}, where \textbf{\{reaction\_info\}} is replaced by a dictionary with reaction information including reactants, products and conditions. The pictures of the reaction mechanism will also be handed over to Gemini.

\begin{lstlisting}[frame=single, breaklines=true, basicstyle=\ttfamily\small, literate={"}{{\char34}}1]

You are an expert in organic chemistry reaction mechanisms.

I will provide you with:
1. A reaction description in JSON format
2. A mechanism image showing the step-by-step reaction mechanism

REACTION DATA:
{reaction_info}

TASK: Analyze the mechanism image and generate a detailed mechanism JSON with each elementary step.

CRITICAL RULES:
1. CHARGE CONSERVATION: The total charge of reactants MUST equal the total charge of products in EACH mechanism step.
   - Count charges carefully: [OH-] has charge -1, [H+] has charge +1, neutral molecules have charge 0
   - For each step, verify: sum(reactant charges) = sum(product charges)

2. LABEL CLASSIFICATION:
   - "substrate": The main organic compound at the start of the reaction
   - "reagent": External reagents that participate in the reaction (like [OH-], [H+], etc.)
   - "intermediate": Products of one step that become reactants in the next step
   - "leaving": Molecules that leave the intermediate products (like [CH3O-], H2O, etc.)
   - "product": The final product of the entire mechanism

3. INTERMEDIATE INHERITANCE:
   - The intermediate product of step N becomes the intermediate reactant of step N+1
   - Track the flow of the main organic molecule through all steps
   - The SMILES of an intermediate product must exactly match the intermediate reactant in the next step

4. ELECTRON NUMBER: Calculate total valence electrons for each molecule
   - H: 1, C: 4, N: 5, O: 6, F: 7, Cl: 7, Br: 7, I: 7, S: 6, P: 5
   - Add electrons for negative charges, subtract for positive charges

5. SMILES ACCURACY:
   - Convert all structures to accurate SMILES format
   - Include stereochemistry where applicable
   - Verify that SMILES correctly represents the structure in the image
   - Pay attention to charged species: use [OH-], [H+], [O-], etc.
   - IMPORTANT: For protonated species, put the H inside the bracket with the charge:
     * CORRECT: [OH+] (protonated oxygen/oxonium)
     * WRONG: [O+](H), [O+]H or [O+H]
     * CORRECT: [OH2] (water)
     * WRONG: [H2O]     
     * CORRECT: [NH4+] (ammonium)
     * WRONG: [N+](H)(H)(H)H
     * CORRECT: [OH2+] (hydronium-like)
     * WRONG: [O+](H)H
     * CORRECT: [nH+]1ccccc1 (pyridine-like)
     * WRONG: [nH]1ccccc1+
     * CORRECT: [NH3] (ammonia)
     * WRONG: NH3     
     * CORRECT: [H2O] (water)
     * WRONG: H2O
   - IMPORTANT: For Phosphorus compounds, use the correct SMILES format:
     * CORRECT: ClP(Cl)Cl
     * WRONG: PCl3
   - IMPORTANT: For hydride ions like [H-], use the correct SMILES format:
     * CORRECT: [HgH]
     * WRONG: [Hg]H
   - IMPORTANT: For metal-monoolefin ternary cyclic complex compounds, use the correct SMILES format:
     * CORRECT: C1=C[Hg+]1
     * WRONG: C1=[C+]1[Hg]
   - IMPORTANT: For carbanions ([CH-], [C-]), write them INLINE in the chain, NOT as a branch:
     * CORRECT: COC(=O)C[CH-]C(=O)OC (carbanion is part of the main chain)
     * WRONG: COC(=O)CC([CH-])C(=O)OC (carbanion as branch adds extra carbon!)
     * CORRECT: CC[CH-]CC (carbanion inline)
     * WRONG: CCC([CH-])C (carbanion as branch - this adds an extra carbon atom)
     * The [CH-] or [C-] must replace a carbon in the chain, not be attached as a substituent
   - IMPORTANT: For alpha-carbonyl anions, prefer ENOLATE form over ketone carbanion form:
     * CORRECT: CC(=C[O-])OC (enolate anion - negative charge on oxygen)
     * WRONG: CC(=O)[CH-]OC (carbanion form - negative charge on carbon)
     * CORRECT: C/C([O-])=C/C (enolate with defined stereochemistry)
     * WRONG: CC(=O)[CH-]C (ketone carbanion)
     * Enolates are more stable due to resonance delocalization of the negative charge
     * When deprotonation occurs alpha to a carbonyl, draw the enolate tautomer

6. REACTION_STEP NUMBERING: Assign step numbers based on reagent addition batches
   - Conditions added in the SAME batch (same time/same arrow in the mechanism) get the SAME step number
   - The conditions in the input JSON are already ordered from first to last batch
   - Look at the mechanism image to identify separate addition steps (usually separated by arrows)
   - Example: If "H2SO4", "EtOH", "reflux" are added together first $\to$ all get reaction_step="1"
   - Example: If "NaOH", "heat" are added in a second step $\to$ both get reaction_step="2"
   - The step numbers should be "1", "2", "3"... as strings, corresponding to each batch of additions

OUTPUT FORMAT - Return a JSON object with this exact structure:
{{
    "reactions": [
        {{
            "reaction_id": "1",
            "reactants": [
                {{
                    "smiles": "REACTANT_SMILES",
                    "label": "None"
                }}
            ],
            "conditions": [
                {{
                    "role": "reagent|solvent|temperature|catalyst",
                    "text": "condition_text",
                    "reaction_step": "step_number"
                }}
            ],
            "products": [
                {{
                    "smiles": "PRODUCT_SMILES",
                    "label": "None"
                }}
            ]
        }}
    ],
    "mechanism": [
        {{
            "mechanism_id": "1",
            "reactants": [
                {{
                    "smiles": "SMILES",
                    "label": "substrate|reagent|intermediate",
                    "electron_number": NUMBER,
                    "charge": NUMBER
                }}
            ],
            "products": [
                {{
                    "smiles": "SMILES",
                    "label": "intermediate|leaving|product",
                    "electron_number": NUMBER,
                    "charge": NUMBER
                }}
            ]
        }},
        {{
            "mechanism_id": "2",
            "reactants": [
                {{
                    "smiles": "INTERMEDIATE_FROM_STEP_1",
                    "label": "intermediate",
                    "electron_number": NUMBER,
                    "charge": NUMBER
                }}
            ],
            "products": [
                {{
                    "smiles": "SMILES",
                    "label": "intermediate|leaving|product",
                    "electron_number": NUMBER,
                    "charge": NUMBER
                }}
            ]
        }}
    ]
}}

VALIDATION CHECKLIST (verify before returning):
1. [ ] Each mechanism step has balanced charges (sum of reactant charges = sum of product charges)
2. [ ] Intermediate products correctly become intermediate reactants in the next step
3. [ ] All SMILES are valid and represent the correct structures
4. [ ] Electron numbers are correctly calculated
5. [ ] Labels are correctly assigned (substrate, reagent, intermediate, leaving, product)
6. [ ] The first mechanism step starts with the substrate
7. [ ] The last mechanism step produces the final product
8. [ ] Conditions are properly assigned to reaction steps

Return ONLY the JSON object, nothing else.

\end{lstlisting}

\paragraph{The prompt of Gemini annotation for training datasets}, given full trajectories:

\begin{lstlisting}[frame=single, breaklines=true, basicstyle=\ttfamily\small, literate={"}{{\char34}}1]

pathway_annotator_system_prompt: |
  # Role
  You are an expert Organic Chemistry Reasoning Engine generating training data for a student model.

  # Task
  Given a complete reaction trajectory (starting reactants -> intermediates -> final product),
  generate predictive chain-of-thought reasoning for each step.

  **Critical**: Write reasoning as if you are predicting forward, NOT explaining a known answer.
  - Use predictive language: "will attack", "is expected to", "should form"
  - Avoid hindsight phrases: "as shown in the product", "the result confirms"

  # Reasoning Guidelines

  For each step, analyze:
  1. Current reactive species and their electronic properties
  2. Which interaction will occur and why
  3. The electron flow mechanism (if applicable)
  4. The resulting intermediate structure

  **Granularity**:
  - **Be concise** for chemically obvious steps:
    - Proton transfer to/from a strong acid/base
    - Unambiguous nucleophilic attack (single electrophilic site)
    - Leaving group departure with clear driving force

  - **Be detailed** when the step involves:
    - Selectivity choices (regio-, stereo-, chemo-)
    - Competing pathways (SN1 vs SN2, E1 vs E2)
    - Non-obvious electronic effects
    - Rearrangements or unusual mechanisms

  # Atom ID Constraint

  **CRITICAL**: Before writing reasoning, mentally audit the atom IDs present in the SMILES.
  Use ONLY atom IDs that exist in the current step's reactants.
  Format: [Element:ID] (e.g., [C:3], [O:18]).
  Do NOT invent or hallucinate atom IDs.

  # Output Format

  ## Reasoning

  **Step 1:**
  [Analysis in forward-prediction style. End with a brief statement of the predicted outcome.]

  **Step 2:**
  [Analysis...]

  ...

  **Step N (Terminal):**
  [Analysis of the final step. Explain why this is the stable endpoint.]

  ## Result

  ```json
  [
    {
      "step_id": 1,
      "reaction_type": "Reaction class name (e.g., Nucleophilic Addition, Proton Transfer, Elimination)",
      "description": "Brief description of what happens",
      "product_smiles": "[Copy EXACTLY from ground truth]",
      "reactive_sites": {
        "nucleophile": "[X:n] or null",
        "electrophile": "[Y:m] or null",
        "leaving_group": "[Z:k] or null"
      },
      "is_terminal": false
    },
    ...
    {
      "step_id": N,
      "reaction_type": "...",
      "description": "...",
      "product_smiles": "...",
      "reactive_sites": {...},
      "is_terminal": true
    }
  ]
  ```

pathway_annotator_template_prompt: |
  **Task:** Generate predictive reasoning for each step of this multi-step reaction.

  **Starting Reactants:**
  {starting_reactants}

  **Reaction Conditions:**
  {conditions}

  **Reaction Trajectory (Ground Truth):**
  {trajectory}

  Generate the Reasoning and Result sections following the system instructions.

\end{lstlisting}

\paragraph{The prompt of checkpoint generation}:
\label{para:ckpt}

\begin{lstlisting}[frame=single, breaklines=true, basicstyle=\ttfamily\small, literate={"}{{\char34}}1]

You are an expert in organic chemistry reaction mechanisms.

I will provide you with a detailed mechanism JSON file containing elementary reaction steps.
Your task is to identify KEY CHECKPOINTS - the critical intermediates that represent major transformations.

MECHANISM DATA:
{mechanism_info}

CHECKPOINT SELECTION CRITERIA:

1. BOND TOPOLOGY CHANGE (Most Critical):
   - Any step where C-C, C-N, or C-O bonds are FORMED or BROKEN
   - Examples: nucleophilic addition, elimination, substitution, cyclization, ring-opening
   - This is the PRIMARY criterion for checkpoint selection

2. FUNCTIONAL GROUP TRANSFORMATION:
   - When the molecule converts from one compound class to another
   - Examples: aldehyde -> imine, imine -> nitrile, nitrile -> amide, alcohol -> ether
   - Record the representative structure at each transformation point

3. COLLAPSE TRIVIAL PROTON TRANSFERS:
   - Simple protonation/deprotonation steps that don't change the carbon skeleton
   - Consecutive proton gain/loss steps should be merged into the SAME checkpoint
   - Only the FINAL intermediate after proton equilibration should be the checkpoint
   - Exception: If protonation is essential for activating a leaving group, it should be included

RULES FOR CHECKPOINT GROUPING:
- If multiple mechanism_ids represent the SAME key intermediate (just different protonation states), 
  group them on the SAME line separated by commas
- For example, if step 5, 6, 7 all represent variants of an imine intermediate with different protonation,
  output: "5,6,7" on one line (any of these appearing counts as correct)
- Each line should represent ONE conceptual checkpoint

ANALYSIS PROCESS:
1. Identify all steps with C-C, C-N, C-O bond changes
2. Identify functional group transformations
3. Group consecutive proton transfer steps
4. Determine which mechanism_ids should be checkpoints

OUTPUT FORMAT:
Return ONLY a simple text format with one checkpoint per line:
- Each line contains mechanism_id(s) that represent that checkpoint
- If multiple mechanism_ids are equivalent, separate them with commas (no spaces)
- Order the lines from first checkpoint to last
- Do not include the starting material or final product (only intermediates)

EXAMPLE OUTPUT:
1
5,6,7
8
12

This means:
- Step 1 is the first checkpoint
- Steps 5, 6, or 7 are equivalent representations of the second checkpoint
- Step 8 is the third checkpoint
- Step 12 is the fourth checkpoint

Return ONLY the checkpoint list, nothing else. No explanations, no formatting, just the lines.

\end{lstlisting}

\clearpage
\newpage

\section{Case Studies}
\label{app:case}

\paragraph{LLMs Captures Textbook Reactivity Trends}

Reactivity trends and heuristics, often derived from fundamental principles and corroborated by experimental results, play a key role in teaching and understanding reactivity. These principles enable chemists to reason bottom-up, starting from basic axioms to deduce likely reaction outcomes. In contrast, current large language models (LLMs) often rely on top-down statistical learning, frequently mapping reagents to products based on memorized "name reaction" patterns rather than physical chemistry. Therefore, a natural question arises: do the same fundamental principles emerge from models trained on large-scale chemical corpora? Do such models reflect the underlying chemical reasoning encoded in human-developed heuristics when facing complex, condition-responsive scenarios? To address this, we scrutinize the step-wise predictions of various state-of-the-art models, examining whether they align with well-established textbook principles across a spectrum of challenging chemical environments.

\begin{figure*}[!]
    \centering
    \includegraphics[width=1\linewidth]{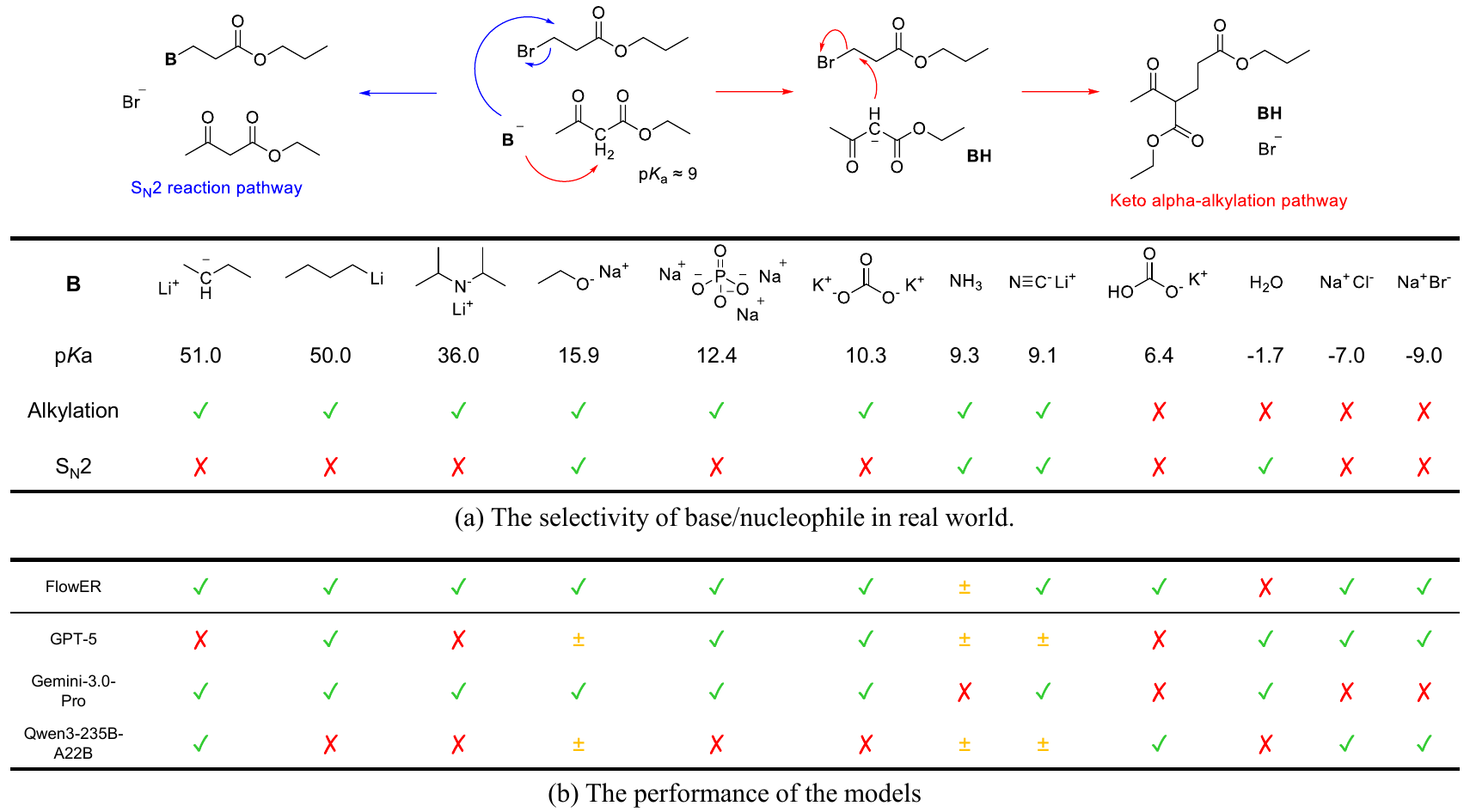}
    \caption{Evaluating model predictions on $pK_a$-dependent selectivity between $S_N2$ and keto alpha-alkylation pathways. }
    \label{fig:s1}
\end{figure*}

Many organic reactions are strongly influenced by the delicate balance between basicity and nucleophilicity. As a textbook-level description, whether a reaction proceeds via proton abstraction (acting as a base) or direct substitution (acting as a nucleophile) heavily depends on the $pK_a$ values and the steric profiles of the species involved. To probe this, we designed a benchmark evaluating the competition between keto alpha-alkylation (requiring a strong base) and the $S_N2$ reaction pathway (requiring a good nucleophile) across a wide continuum of reagents, ranging from strong bases ($pK_a \approx 51.0$) to weak conjugate bases ($pK_a \approx -9.0$) (fig.~\ref{fig:s1}a). By tracking how different models navigate this gradient, we can assess their data-driven understanding of $pK_a$-dependent reactivity (fig.~\ref{fig:s1}b).

\begin{figure*}
    \centering
    \includegraphics[width=1\linewidth]{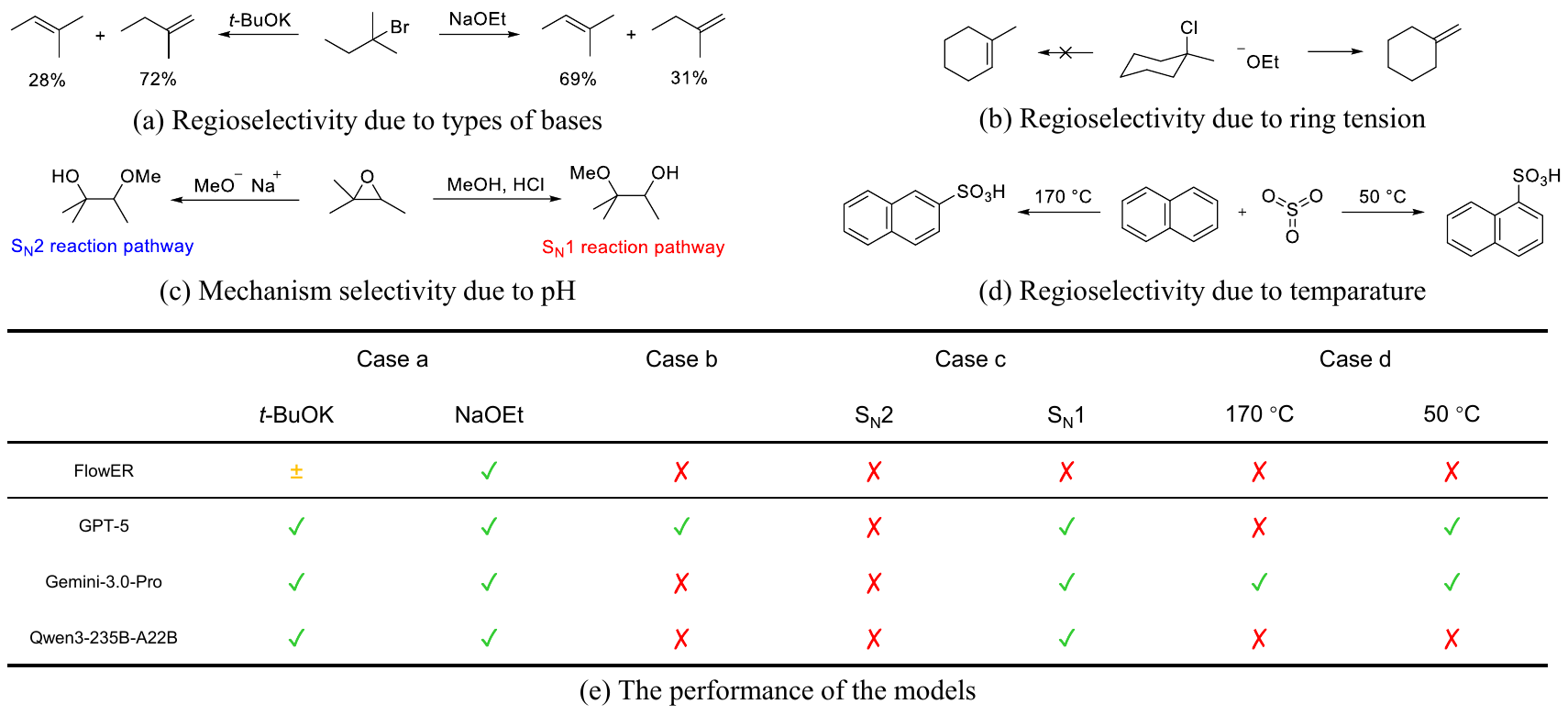}
    \caption{Evaluating model performance on complex, condition-responsive regioselectivity and mechanistic divergence.}
    \label{fig:s2}
\end{figure*}

Beyond simple acid-base chemistry, true chemical intelligence requires the ability to predict condition-responsive regioselectivity and mechanism divergence. In fig.~\ref{fig:s2}, we further challenge the models with classic mechanistic dilemmas: (a) regioselectivity dictated by the steric bulk of the base, such as the bulky t-BuOK favoring Hofmann elimination versus the smaller NaOEt favoring Zaitsev elimination; (b) regioselectivity driven by intrinsic ring tension; (c) pH-dependent mechanistic shifts between $S_N1$ and $S_N2$ pathways; and (d) thermodynamic versus kinetic control driven by temperature gradients, such as shifts between 170$^{\circ}$C and 50$^{\circ}$C.

As shown in the evaluation tables (fig.~\ref{fig:s1}b and fig.~\ref{fig:s2}e), small specialized generative models, such as FlowER, often exhibit rigid predictive behaviors. Because they heavily rely on the in-domain statistical distribution of their training data, they frequently fail to adapt when reaction conditions or reagent $pK_a$ values shift. In contrast, state-of-the-art LLMs—including GPT-5, Gemini-3.0-Pro, and models from the Qwen3 family—demonstrate a distinctly higher degree of generalized chemical intuition. While they may not perfectly execute these physical chemistry rules in every single instance, LLMs show a discernible, emergent capacity to recognize the impact of steric bulk, temperature gradients, and $pK_a$ thresholds. This suggests that the broad linguistic and scientific data encountered during large-scale pre-training endows LLMs with a more robust foundation for condition-responsive chemical reasoning than smaller, purely dataset-driven models.

\clearpage
\newpage

\paragraph{FukuyamaBench Case Studies} 

\begin{figure}[t]
    \centering
    \includegraphics[width=1\linewidth]{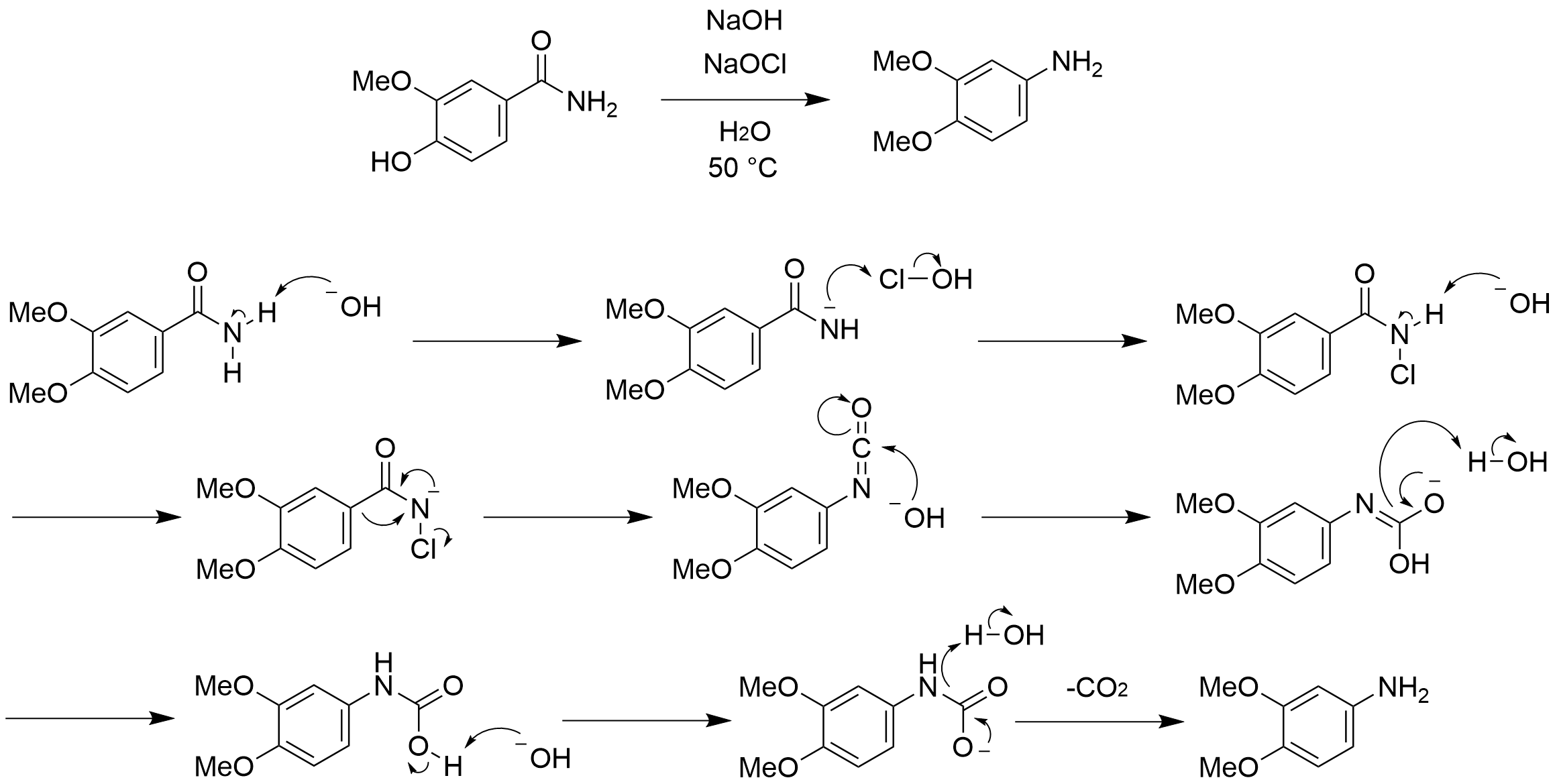}
    \caption{Full reaction mechanism of FukuyamaBench A057.}
    \label{fig:s3}
\end{figure}

The full reaction mechanism of fig.~\ref{fig:4} is shown in fig.~\ref{fig:s3} and ~\ref{fig:s4}. 

\textbf{Case A057}

The case follows the Hofmann-style sequence: N-chlorination of the amide, deprotonation to N-chloroamide anion, elimination to isocyanate, hydration/proton transfer to carbamic acid, then decarboxylation. Our model identifies the correct mechanistic backbone: chlorination is followed by Hofmann rearrangement to the isocyanate, and only then hydration/decarboxylation. This keeps the trajectory anchored on the chemically meaningful intermediates used as benchmark checkpoints.
    
GPT-5 partially recovers the right scaffold, but it over-specifies spectator species and pushes the sequence into overly explicit ionic bookkeeping. In its best sample, it reaches the N-chloroamide anion, but then represents later states with extra waters/hydroxide/chloride and reformulates the post-isocyanate manifold through bulk hydroxide addition rather than the cleaner checkpoint intermediates. As a result, the reasoning does not land on the benchmark's canonical isocyanate/carbamic-acid states even though the broad idea is similar. Qwen3-30B-A3B makes a more fundamental mechanistic error: it reframes the chemistry as direct hydroxide attack/hydrolysis or generic oxidation of the amide carbonyl, rather than a Hofmann rearrangement initiated by N-chlorination. Once it chooses that wrong elementary step, the whole trajectory stays on an amide-hydrolysis manifold and never reaches the isocyanate branch.

\begin{lstlisting}[frame=single, breaklines=true, basicstyle=\ttfamily\small, literate={"}{{\char34}}1]
A057 CoT excerpt (k=2):
Step 2 (matched checkpoint 1):
The amide anion [N-:8] is a strong nucleophile. Sodium hypochlorite (NaOCl) is present in the reaction mixture, providing the hypochlorite ion ([O-]Cl), which acts as an electrophile at the oxygen atom. The amide anion [N-:8] is expected to attack the electrophilic oxygen of the hypochlorite. Simultaneously, the nitrogen atom [N:8] will reform a double bond with the carbonyl carbon [C:7], leading to the expulsion of the chloride ion as a leaving group. This will result in the formation of an N-chloroamide.
Predicted outcome: N-chlorination of the amide.
Generated structure:
[CH3:1][O:2][c:3]1[cH:4][cH:5][c:6]([C:7](=[O:9])[NH:8][Cl:14])[cH:10][c:11]1[O:12][CH3:13]

Step 4 (matched checkpoint 2):
The intermediate contains a negatively charged nitrogen atom [N-:8] adjacent to a carbonyl group [C:7]=[O:9] and a chlorine atom [Cl:14] attached to [N:8]. This structure is poised for a concerted elimination. The lone pair on [N-:8] is expected to collapse to form a triple bond with [C:7], simultaneously causing the C=O pi electrons to move onto [O:9], resulting in the expulsion of the chloride ion. This process will yield a highly reactive isocyanate.
Predicted outcome: Formation of an isocyanate via elimination.
Generated structure:
[CH3:1][O:2][c:3]1[cH:4][cH:5][c:6]([N:8]=[C:7]=[O:9])[cH:10][c:11]1[O:12][CH3:13]

Step 6 (matched checkpoint 3):
The intermediate contains a positively charged oxygen atom [OH2+:14] and a negatively charged nitrogen atom [N-:8]. An intramolecular or intermolecular proton transfer is expected to occur. The negatively charged nitrogen [N-:8] will likely abstract a proton from the positively charged oxygen [OH2+:14], resulting in a neutral carbamic acid and a neutral amine.
Predicted outcome: Proton transfer to yield a carbamic acid and an amine.
Generated structure:
[CH3:1][O:2][c:3]1[cH:4][cH:5][c:6]([NH:8][C:7](=[O:9])[OH:14])[cH:10][c:11]1[O:12][CH3:13]
\end{lstlisting}

\textbf{Case A063}

The case keeps the protonated oxonium intermediate, then performs beta-elimination to form the conjugated product manifold; the matched branch lands on the benchmark checkpoint product. Our model treats protonation as activation rather than as the endpoint. After forming the oxonium, it continues along the productive elimination channel that generates the conjugated enol-ether intermediate and then the neutral rearranged product manifold, which is exactly what the benchmark is checking.

GPT-5 stops too early conceptually: it treats oxonium formation as the terminal answer. This means the model recognizes acid activation but never executes the crucial downstream C--O cleavage/elimination sequence, so it matches the first checkpoint only and misses the actual transformation-driving step. Qwen3-30B-A3B does continue beyond protonation, but it follows the wrong continuation logic. Instead of committing to the benchmark's elimination pathway, it invents carbocation, hydride-shift, or ring-opening style branches that are not the productive route here. In other words, the failure is not merely missing a checkpoint; it is choosing the wrong post-activation mechanistic manifold.

\begin{lstlisting}[frame=single, breaklines=true, basicstyle=\ttfamily\small, literate={"}{{\char34}}1]
A063 CoT excerpt (k=3; lightly reformatted):

Step 1 (matched checkpoint 1):
The starting material contains a cyclic acetal-like structure with two potential ether oxygen atoms, [O:4] and [O:6]. In the presence of a strong acid like TsOH, the most basic site, an ether oxygen, will be protonated. Both [O:4] and [O:6] are equivalent in terms of basicity. Protonation of [O:6] will activate the adjacent [O:6]-[CH2:7] bond for cleavage, making [CH2:7] a better leaving group (as a neutral alkene). The acid will abstract a proton from [O:6], forming a protonated ether.
Predicted outcome: Protonation of the cyclic ether oxygen.
Generated structure:
[CH2:1]=[CH:2][CH2:3][O:4][C:5]1([OH+:6][CH2:7][CH:8]=[CH2:9])[CH2:10][CH2:11][CH2:12][CH2:13][CH2:14]1

Step 3 (matched checkpoint 2):
The carbocation at [C:5] is a strong electrophile. An elimination reaction is expected to occur to relieve the positive charge and form a more stable, conjugated system. The most acidic protons are located at [CH2:10], which are beta to the carbocation. A base (likely a solvent molecule or the conjugate base of TsOH) will abstract a proton from [CH2:10], and the electrons from the C-[H] bond will collapse to form a new pi bond between [C:5] and [CH2:10]. This will result in a conjugated dienone system.
Predicted outcome: Deprotonation at [CH2:10] to form a conjugated dienone.
Generated structure:
[CH2:1]=[CH:2][CH2:3][O:4][C:5]1=[CH:10][CH2:11][CH2:12][CH2:13][CH2:14]1
\end{lstlisting}

\begin{figure}[!h]
    \centering
    \includegraphics[width=1\linewidth]{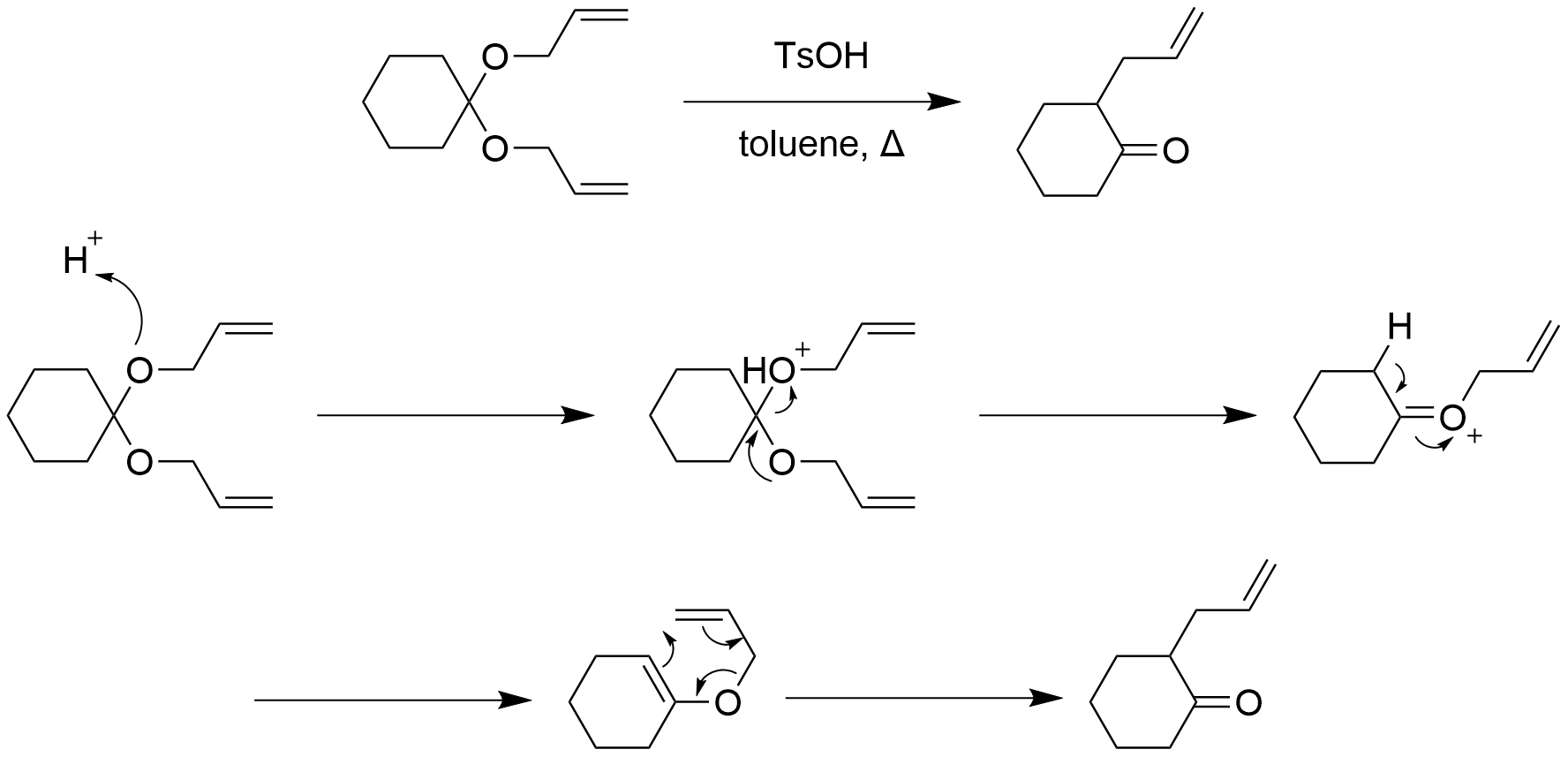}
    \caption{Full reaction mechanism of FukuyamaBench A063.}
    \label{fig:s4}
\end{figure}

\end{document}